\documentclass[preprint,11pt]{elsarticle}
\usepackage[latin1]{inputenc}
\usepackage[english]{babel}
\usepackage{amssymb,amsmath,graphicx}
\usepackage{multirow}
\usepackage{natbib}
\usepackage{enumitem}
\usepackage{setspace}
\usepackage[table]{xcolor}
\usepackage{microtype}
\usepackage{varwidth}
\usepackage{afterpage,lscape}
\newcolumntype{M}[1]{>{\centering\let\newline\\\arraybackslash\hspace{0pt}}m{#1}}
\usepackage{hhline}
\usepackage{nicefrac}
\usepackage[norelsize,boxed,linesnumbered,noend]{algorithm2e}
\SetArgSty{textrm} 
\SetAlCapSkip{0.7em} 
\usepackage[hidelinks]{hyperref}
\usepackage{multicol}
\usepackage[FIGTOPCAP]{subfigure}
\usepackage{caption}

\renewcommand{\Re}{{\mathbb{R}}}

\newcommand{\cO}{{\cal O}}

\newcolumntype{P}[1]{>{\centering}p{#1}}
\newcolumntype{M}[1]{>{\centering}m{#1}}
\newcolumntype{H}{>{\setbox0=\hbox\bgroup}c<{\egroup}@{}}
\definecolor{light-gray}{gray}{0.95}
\newcommand{\norm}[1]{\left\lVert#1\right\rVert}
\DeclareMathOperator*{\argmin}{arg\,min}

\newcommand{\myblue}[1]{#1}

\newtheorem{property}{Property}

\DeclareGraphicsExtensions{.pdf, .png}

\allowdisplaybreaks
\title{HG-means: A scalable hybrid genetic algorithm for minimum sum-of-squares clustering}

\author{Daniel Gribel, Thibaut Vidal}

\begin{document}

\linespread{1.0}\selectfont

\vspace*{-1.8cm}

\begin{scriptsize}
\noindent
This is the post-peer-review, pre-copyedit version of the article published in \emph{Pattern Recognition}. The published article is available at \url{http://dx.doi.org/10.1016/j.patcog.2018.12.022}. This manuscript version is made available under the CC-BY-NC-ND 4.0 license \url{http://creativecommons.org/licenses/by-nc-nd/4.0/}.\par
\end{scriptsize}

\vspace*{0.8cm}

\begin{center}

\vspace*{-0.5cm}

\begin{huge}
\textsc{HG-means}: A scalable hybrid genetic algorithm \\ \vspace*{0.3cm} for minimum sum-of-squares clustering
\end{huge}

\vspace*{0.6cm}

\textbf{Daniel Gribel$^a$, Thibaut Vidal$^{a*}$} \\
$^a$ Departamento de Inform\'{a}tica, Pontif\'{i}cia Universidade Cat\'{o}lica do Rio de Janeiro (PUC-Rio),  Rua Marqu\^{e}s de S\~{a}o Vicente, 225 - G\'{a}vea, Rio de Janeiro - RJ, 22451-900, Brazil \\
\{vidalt,dgribel\}@inf.puc-rio.br \\
\vspace*{0.15cm}

\end{center}
\noindent
\textbf{Abstract.}
Minimum sum-of-squares clustering (MSSC) is a widely used clustering model, of which the popular \textsc{K-means} algorithm constitutes a local minimizer. It is well known that the solutions of \textsc{K-means} can be arbitrarily distant from the true MSSC global optimum, and dozens of alternative heuristics have been proposed for this problem. However, no other algorithm has been predominantly adopted in the literature. This may be related to differences of computational effort, or to the assumption that a near-optimal solution of the MSSC has only a marginal impact on \myblue{clustering} validity.

In this article, we dispute this belief.
We introduce an efficient population-based metaheuristic that uses \textsc{K-means} as a local search in combination with problem-tailored crossover, mutation, and diversification operators. This algorithm can be interpreted as a multi-start \textsc{K-means}, in which the initial center positions are carefully sampled based on the search history.
The approach is scalable and accurate, outperforming all recent state-of-the-art algorithms for MSSC in terms of solution quality, measured by the depth of local minima. This enhanced accuracy leads to \myblue{clusters} which are  significantly closer to the ground truth than those of other algorithms, for overlapping Gaussian-mixture datasets with a large number of features. Therefore, improved global optimization methods appear to be essential to better exploit the MSSC model in high dimension.
\vspace*{0.2cm}

\vspace*{-0.3cm}

\noindent
$^*$ Corresponding author

\noindent
Declarations of interest: none

\linespread{1.0}\selectfont

\section{Introduction}
\label{Intro}

Broadly defined, clustering is the problem of organizing a collection of elements into coherent groups in such a way that similar elements are in the same cluster and different elements are in different clusters. Of the models and formulations for this problem, the Euclidean minimum sum-of-squares clustering (MSSC) is prominent in the literature. MSSC can be formulated as an optimization problem in which the objective is to minimize the sum-of-squares of the Euclidean distances of the samples to their cluster means. This problem has been extensively studied over the last 50 years, as highlighted by various surveys and books \citep[see, e.g.,][]{Han2011,Hruschka2009,Jain2010}.

The NP-hardness of MSSC \citep{Aloise2009a} and the size of practical datasets explain why most MSSC algorithms are heuristics, designed to produce an approximate solution in a reasonable computational time.
\textsc{K-means}~\citep{Hartigan1979} (also called Lloyd's algorithm \citep{Lloyd1982}) and \textsc{K-means}++~\citep{Arthur2007} are two popular local search algorithms for MSSC that differ in the construction of their initial solutions. Their simplicity and low computational complexity explain their extensive use in practice. However, these methods have two significant disadvantages: (i) their solutions can be distant from the global optimum, especially in the presence of a large number of clusters and dimensions, and (ii) their performance is sensitive to the initial conditions of the search.

To circumvent these issues, a variety of heuristics and metaheuristics have been proposed with the aim of better escaping from shallow local minima (i.e., poor solutions in terms of the MSSC objective). Nearly all the classical metaheuristic frameworks have been applied, including simulated annealing, tabu search, variable neighborhood search, iterated local search, evolutionary algorithms \citep{Al-Sultan1995,Hansen2001a,Ismkhan2018,Maulik2000,Sarkar1997,Selim1991}, as well as more recent incremental methods and convex optimization techniques \citep{HoaiAn2014,Bagirov2016,Bagirov2011,Karmitsa2017}. However, these sophisticated methods have not been predominantly used in machine learning applications. This may be explained by three main factors: 1) data size and computational time restrictions, 2) the limited availability of implementations, or 3) the belief that a near-optimal solution of the MSSC model has little impact on \myblue{clustering validity}.

To help remove these barriers, we introduce a simple and efficient hybrid genetic search for the MSSC called \textsc{HG-means}, and conduct extensive computational analyses to measure the correlation between solution quality (in terms of the MSSC objective) and \myblue{clustering} performance (based on external measures).
Our method combines the improvement capabilities of the \textsc{K-means} algorithm with a problem-tailored crossover, an adaptive mutation scheme, and population-diversity management strategies. The overall method can be seen as a multi-start \textsc{K-means} algorithm, in which the initial center positions are sampled by the genetic operators based on the search history.
\textsc{HG-means}' crossover uses a minimum-cost matching algorithm as a subroutine, with the aim of inheriting genetic material from both parents without excessive perturbation and creating \emph{child} solutions that can be improved in a limited number of iterations. The adaptive mutation operator has been designed to help cover distant samples without being excessively attracted by outliers. Finally, the population is managed so as to prohibit \emph{clones} and favor the discovery of diverse solutions, a feature that helps to avoid premature convergence toward low-quality local minima. 

As demonstrated by our experiments on a variety of datasets, \textsc{HG-means} produces MSSC solutions of significantly higher quality than those provided by previous algorithms. Its computational time is also lower than that of recent state-of-the-art optimization approaches, and it grows linearly with the number of samples and dimension. Moreover, when considering the reconstruction of a mixture of Gaussians, we observe that the standard repeated \textsc{K-means} and \textsc{K-means++} approaches remain trapped in shallow local minima which can be very far from the ground truth, whereas \textsc{HG-means} consistently attains better local optima and \myblue{finds more accurate clusters}. The performance gains are especially pronounced on datasets with a larger number of clusters and a feature space of higher dimension, in which more independent information is available, but also in which pairwise distances are known to become more uniform and less meaningful. Therefore, some key challenges associated to high-dimensional data clustering may be overcome by improving the optimization algorithms, before even considering a change of clustering model or paradigm.

The remainder of this article is structured as follows.
Section \ref{pb-statement} formally defines the MSSC and reviews the related literature.
Section \ref{chap:methodology} describes the proposed \textsc{HG-means} algorithm. Section \ref{chap:experiments} reports our computational experiments, and Section \ref{sec:conclusions} provides some concluding remarks.

\section{Problem Statement}
\label{pb-statement}

In a clustering problem, we are given a set $P = \{p_1,\dots,p_n\}$ of $n$ samples, where each sample $p_i$ is represented as a point in $\mathbb{R}^d$ with coordinates $(p^1_i,\dots,p^d_i)$, and we seek to partition $P$ into $m$ disjoint clusters $\mathcal{C} = (C_1,\dots,C_m)$ so as to minimize a criterion $f(\mathcal{C})$. There is no universal objective suitable for all applications, but $f(\cdot)$ should generally promote homogeneity (similar samples should be in the same cluster) and separation (different samples should be in different clusters). MSSC corresponds to a specific choice of objective function, in which one aims to form the clusters and find a center position $y_k \in \mathbb{R}^d$ for each cluster, in such a way that the sum of the squared Euclidean distances of each point to the center of its associated cluster is minimized. This problem has been the focus of extensive research: there are many applications \citep{Jain2010}, and it is the natural problem for which \textsc{K-means} finds a local minimum.
 
A compact mathematical formulation of MSSC is presented in Equations (\ref{f-obj})--(\ref{f-eq3}).
For each sample and cluster, the binary variable $x_{ik}$ takes the value $1$ if sample $i$ is assigned to cluster $k$, and $0$ otherwise. The variables $y_k \in \mathbb{R}^d$ represent the positions of the centers.
\begin{align}
\min \hspace*{0.4cm} & \sum_{i=1}^{n} \sum_{k=1}^m x_{ik} \norm{p_i - y_k}^2 \label{f-obj} \\
\text{s.t.} \hspace*{0.4cm} & \sum_{k=1}^{m} x_{ik} = 1 & i \in \{1,\dots,n\} \label{f-eq1} \\
& x_{ik} \in \{0,1\} & i \in \{1,\dots,n\}, k \in \{1,\dots,m\} \label{f-eq2} \\
& y_k \in \mathbb{R}^d & k \in \{1,\dots,m\} \label{f-eq3}
\end{align}
In the objective, $\norm{\cdot}$ represents the Euclidean norm. Equation (\ref{f-eq1}) forces each sample to be associated with a cluster, and Equations (\ref{f-eq2})--(\ref{f-eq3}) define the domains of the variables.
Note that in this model, and in the remainder of this paper, we consider a fixed number of clusters $m$. Indeed, from the MSSC objective viewpoint, it is always beneficial to use the maximum number of available clusters. For some applications such as color quantization and data compression \citep{Scheunders1997a}, the number of clusters is known in advance (desired number of colors or compression factor). Analytical techniques have been developed to find a suitable number of clusters \citep{Sugar2003} when this information is not available. Finally, it is common to solve MSSC for a range of values of $m$ and select the most relevant result a-posteriori.

Regarding computational complexity, MSSC can be solved in $\mathcal{O}(n^3)$ time when $d=1$ using dynamic programming. For general $m$ and $d$, MSSC is NP-hard \citep{Aloise2009a,Aloise2012exact}.
Optimal MSSC solutions are known to satisfy at least two necessary conditions: 

\begin{property}
\linespread{1.0}
\emph{In any optimal MSSC solution, for each $k \in \{1,\dots,m\}$, the position of the center~$y_k$ coincides with the centroid of the points belonging to $C_k$:
\begin{equation}
y_k = \frac{1}{|C_k|} \sum_{i \in C_k} p_i.
\end{equation}
}
\end{property}

\begin{property}
\linespread{1.0}
\emph{In any optimal MSSC solution, for each $i \in \{1,\dots,n\}$, the sample $p_i$ is associated with the closest cluster $C_{k_\textsc{min}(i)}$ such that: 
\begin{equation}
k_\textsc{min}(i) = \argmin\nolimits_{k=1}^n \norm{p_i - y_k}^2.
\end{equation}
}
\end{property}

These two properties are fundamental to understand the behavior of various MSSC algorithms. The \textsc{K-means} algorithm, in particular, iteratively modifies an incumbent solution to satisfy first Property~1 and then Property 2, until both are satisfied simultaneously. Various studies have proposed more efficient data structures and speed-up techniques for this method.
For example, \citet{Hamerly2010} provides an efficient \textsc{K-means} algorithm that has a complexity of $\mathcal{O}(nmd + md^2)$ per iteration. This algorithm is faster in practice than its theoretical worst case, since it avoids many of the innermost loops of \textsc{K-means}.

Other improvements of \textsc{K-means} have focused on improving the choice of initial centers~\citep{Steinley2006}. \textsc{K-means}++ is one such method. This algorithm places the first center $y_1$ in the location of a random sample selected with uniform probability. Then, each subsequent center $y_k$ is randomly placed in the location of a sample $p_j$, with a probability proportional to the distance of $p_j$ to its closest center in $\{y_1,\dots,y_{k-1}\}$. Finally, \textsc{K-means} is executed from this starting point. With this rule, the expected solution quality is within a factor $8 (\log m + 2)$ of the global optimum.

Numerous other solution techniques have been proposed for MSSC. These algorithms can generally be classified according to whether they are exact or heuristic, deterministic or probabilistic, and hierarchical or partitional. Some process complete solutions whereas others construct solutions during the search, and some maintain a single candidate solution whereas others work with population of solutions~\citep{Jain2010}. The range of methods includes construction methods and local searches, metaheuristics, and mathematical programming techniques. Since this work focuses on the development of a hybrid genetic algorithm with population management, the remainder of this review focuses on other evolutionary methods for this problem, adaptations of \textsc{K-means}, as well as algorithms that currently report the best known solutions for the MSSC objective since these methods will be used for comparison purposes in Section~\ref{chap:experiments}.

\citet{Hruschka2009} provide a comprehensive review and analysis of evolutionary algorithms for MSSC, comparing different solution encoding, crossover, and mutation strategies. As is the case for other combinatorial optimization problems, many genetic algorithms do not rely on random mutation and crossover only, but also integrate a local search to stimulate the discovery of high-quality solutions. Such algorithms are usually called \emph{hybrid genetic} or \emph{memetic} algorithms \citep{Blum2011}. The algorithms of \cite{Franti1997,Kivijarvi2003,Krishna1999,Lu2004a,Merz2002} are representative of this type of strategy and exploit \textsc{K-means} for solution improvement. In particular, \cite{Franti1997} and \cite{Kivijarvi2003} propose a hybrid genetic algorithm based on a portfolio of six crossover operators. One of these, which inspired the crossover of the present work, pairs the centroids of two solutions via a greedy nearest-neighbor algorithm and  randomly selects one center from each pair. The mutation operator relocates a random centroid in the location of a random sample, with a small probability. Although this method has some common mechanisms with \textsc{HG-means}, it also misses other key components: an exact matching-based crossover, population-management mechanisms favoring the removal of \emph{clones}, and an adaptive parameter to control the attractiveness of outliers in the mutation. 
The variation (crossover and mutation) operators of \cite{Krishna1999,Lu2004a,Merz2002} are also different from those of \textsc{HG-means}. In particular, \cite{Krishna1999,Lu2004a} do not rely on crossover but exploit random mutation to reassign data points to clusters. Finally, \cite{Merz2002} considers an application of clustering for gene expression data, using \textsc{K-means} as a local search along with a crossover operator that relies on distance proximity to exchange centers between solutions.

Besides evolutionary algorithms and metaheuristics, substantial research has been conducted on incremental variants of the \textsc{K-means} algorithm \citep{Bagirov2008, Bagirov2016, Karmitsa2017, Likas2003, Ordin2015}, leading to the current state-of-the-art results for large-scale datasets. Incremental clustering algorithms construct a solution of MSSC iteratively, adding one center at a time. The global \textsc{K-means} algorithm \citep{Likas2003} is such a method. Starting from a solution with $k$ centers, the complete algorithm performs $n$ runs of \textsc{K-means}, one from each initial solution containing the $k$ existing centers plus sample $i \in \{1,\dots,n\}$. The best solution with $k+1$ centers is stored, and the process is repeated until a desired number of clusters is attained. Faster versions of this algorithm can be designed, by greedily selecting a smaller subset of solutions for improvement at each step. For example, the modified global \textsc{K-means} (MGKM) of \citep{Bagirov2008} solves an auxiliary clustering problem to select one good initial solution at each step instead of considering all $n$ possibilities. This algorithm was improved in \cite{Ordin2015} into a multi-start modified global \textsc{K-means} (MS-MGKM) algorithm, which generates several candidates at each step. Experiments on 16 real-world datasets show that MS-MGKM produces more accurate solutions than MGKM and the global \textsc{K-means} algorithm. These methods were also extended in \cite{Bagirov2016} and \cite{Karmitsa2017}, by solving an optimization problem over a difference of convex (DC) functions in order to choose candidate initial solutions. Finally, \cite{Karmitsa2018} introduced an incremental nonsmooth optimization algorithm based on a limited memory bundle method, which produces solutions in a short time. To this date, the \textsc{MS-MGKM}, \textsc{DCClust} and \textsc{DCD-Bundle} algorithms represent the current state-of-the-art in terms of solution quality. 
 
Despite this extensive research, producing high-quality MSSC solutions in a consistent manner remains challenging for large datasets. Our algorithm, presented in the next section, helps to fill~this~gap.

\section{Proposed Methodology}
\label{chap:methodology}

\textsc{HG-means} is a hybrid metaheuristic that combines the exploration capabilities of a genetic algorithm and the improvement capabilities of a local search, along with general population-management strategies to preserve the diversity of the genetic information. 
Similarly to \cite{Kivijarvi2003,Krishna1999} and some subsequent studies, the \textsc{K-means} algorithm is used as a local search. Moreover, the proposed method differs from previous work in its main variation operators: it relies on an exact bipartite matching crossover, uses a sophisticated adaptive mechanism in the mutation operator, as well as population-diversity management techniques.

The general scheme is given in Algorithm \ref{genetic-algo}.
Each individual $P$ in the population is represented as a triplet $( \psi_P,\phi_P,\alpha_P )$ containing a membership chromosome~$\psi_P$ and a coordinate chromosome~$\phi_P$ to represent the solution and a mutation parameter $\alpha_P$ to help balance the influence of outliers. The algorithm first generates a randomized initial population (Section \ref{sec:solution-representation}) and then iteratively applies variation operators (selection, recombination, mutation) and local search (\textsc{K-means}) to evolve this population. At each iteration, two parents $P_1$ and $P_2$ are selected and crossed (Section \ref{subsec:select-cross}), yielding the coordinates $\phi_C$ and mutation parameter~$\alpha_C$ of an offspring~$C$. A mutation operator is then applied to $\phi_C$ (Section \ref{subsec:mutation}), leading to an individual that is improved using the \textsc{K-means} algorithm (Section \ref{subsec:local-improvement}) and then included in the population.

\begin{algorithm}[htbp]
\linespread{1.0}\selectfont
Initialize population with $\Pi_\textsc{max}$ individuals/solutions \\
\While{(number of iterations without improvement $< N_1$) $\wedge$ (number of iterations $< N_2$)}
	{
		Select parents $P_1$ and $P_2$ by binary tournament \\
		Apply crossover to $P_1$ and $P_2$ to generate an offspring $C$ \\
		Mutate $C$ to obtain $C'$ \\
		Apply local search (\textsc{K-means}) to $C'$ to obtain an individual $C''$ \\
		Add $C''$ to the population \\
		\If{the size of the population exceeds $\Pi_\textsc{max}$}
		{
			Eliminate clones and select $\Pi_\textsc{min}$ survivors \\
		}
}
Return best solution
 \caption{\textsc{HG-means} -- general structure} \label{genetic-algo}
\end{algorithm}

Finally, each time the population exceeds a prescribed size $\Pi_\textsc{max}$, a survivor selection phase is triggered (Section \ref{sec:population-management}), to retain only a diverse subset of $\Pi_\textsc{min}$ good individuals. The algorithm terminates after $N_1$ consecutive iterations (generation of new individuals) without improvement of the best solution or a total of $N_2$ iterations have been performed. The remainder of this section describes each component of the method, in more detail.

\subsection{Solution Representation and Initial Population}
\label{sec:solution-representation}

\begin{figure}[htbp]
\includegraphics[width=\textwidth]{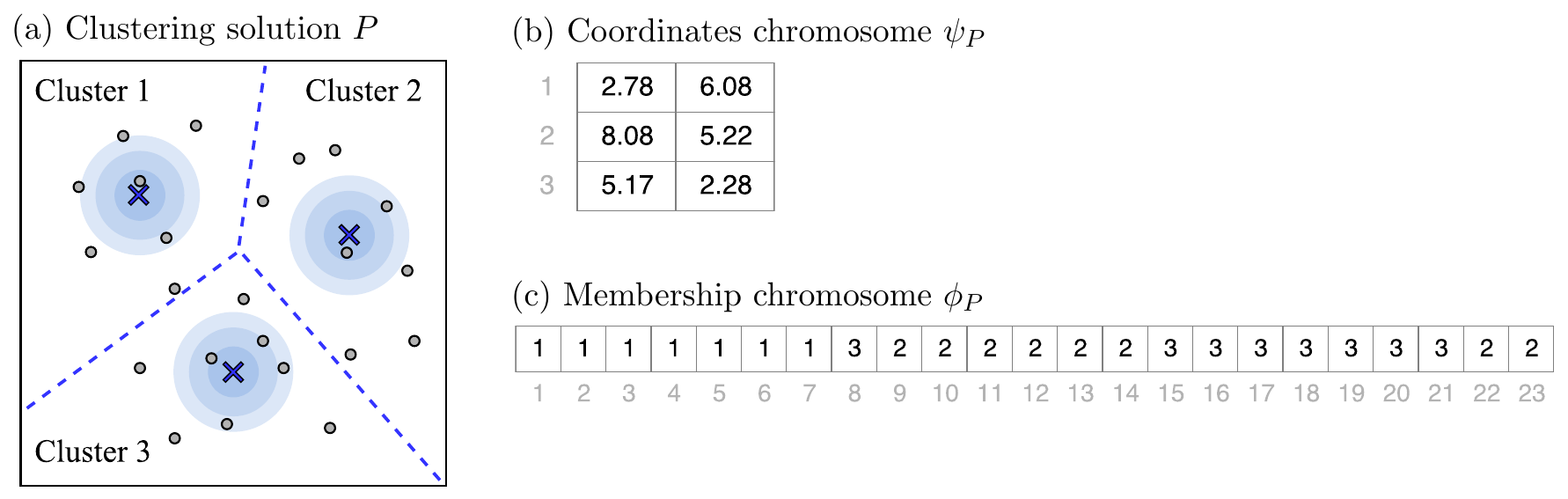}
\caption{Representation of an MSSC solution as a chromosome pair.}
\label{fig:encoding}
\end{figure}

Each individual $P$ contains two chromosomes encoding the solution: a \emph{membership} chromosome $\phi_P$ with $n$ integers, specifying for each sample the index of the cluster with which it is associated; and a \emph{coordinate} chromosome $\psi_P$ with $m$ real vectors in $\Re^d$, representing the coordinates of the center of each cluster. The individual is completed with a mutation parameter, $\alpha_P \in \Re$. Figure~\ref{fig:encoding} illustrates this solution representation for a simple two-dimensional example with three centers.

Observe that either of the two chromosomes is sufficient to characterize an MSSC solution. If only the membership chromosome $\phi_P$ is known, then Property~1 states that the center of each cluster should be located at the centroid of the points associated with it, and a trivial calculation gives the coordinates of each centroid in $\cO(n d)$. If only the coordinate chromosome $\psi_P$ is known, then Property~2 states that each sample should be associated with its closest center, and a simple linear search in $\cO(n m d)$, by calculating the distances of all the centers from each sample, gives the membership chromosome.
Finally, note that the iterative decoding of one chromosome into the other, until convergence, is equivalent to the \textsc{K-means} algorithm.

\subsection{Selection and Crossover}
\label{subsec:select-cross}

The generation of each new individual begins with the selection of two parents $P_1$ and $P_2$.
The parent selection is done by binary tournament. A binary tournament selects two random solutions in the population with uniform probability and retains the one with the best fitness. The fitness considered in \textsc{HG-means} is simply the value of the objective function (MSSC cost).

Then, the coordinate chromosomes $\psi_{P_1}$ and $\psi_{P_2}$ of the two parents serve as input to the \emph{matching crossover} (MX), which generates the coordinate chromosome $\psi_{C}$ of an offspring in two steps:

\begin{itemize}[nosep,leftmargin=*]
\item \textsc{Step 1)} The MX solves a bipartite matching problem to pair-up the centers of the two parents. Let $G = (U, V, E)$ be a complete bipartite graph, where the vertex set $U = (u_1,\dots,u_m)$ represents the centers of parent $P_1$, and the vertex set $V = (v_1,\dots,v_m)$ represents the centers of parent $P_2$. Each edge $(u_i,v_j) \in E$, for $i \in \{1,\dots,m\}$ and $j \in \{1,\dots,m\}$ represents a possible association of center $i$ from parent $P_1$ with center $j$ from parent $P_2$. Its cost $c_{ij} = \norm{\psi_{P_1}(i) - \psi_{P_2}(j)}$ is calculated as the Euclidean distance between the two centers. A minimum-cost bipartite matching problem is solved in the graph $G$ using an efficient implementation of the Hungarian algorithm \citep{Kuhn1955}, returning $m$ pairs of centers in $\cO(m^3)$ time.

\item \textsc{Step 2)} For each pair obtained at the previous step, the MX randomly selects one of the two centers with equal probability, leading to a new \emph{coordinate} chromosome with $m$ centers and inherited characteristics from both parents.
\end{itemize}

Finally, the mutation parameter of the offspring is obtained as a simple average of the parent values: $\alpha_C = \frac{1}{2} (\alpha_{P_1} + \alpha_{P_2})$.

\begin{figure}[htb]
\centering
\subfigure[Parent $p_1$]{\label{fig:cr01}\includegraphics[width=0.33\textwidth]{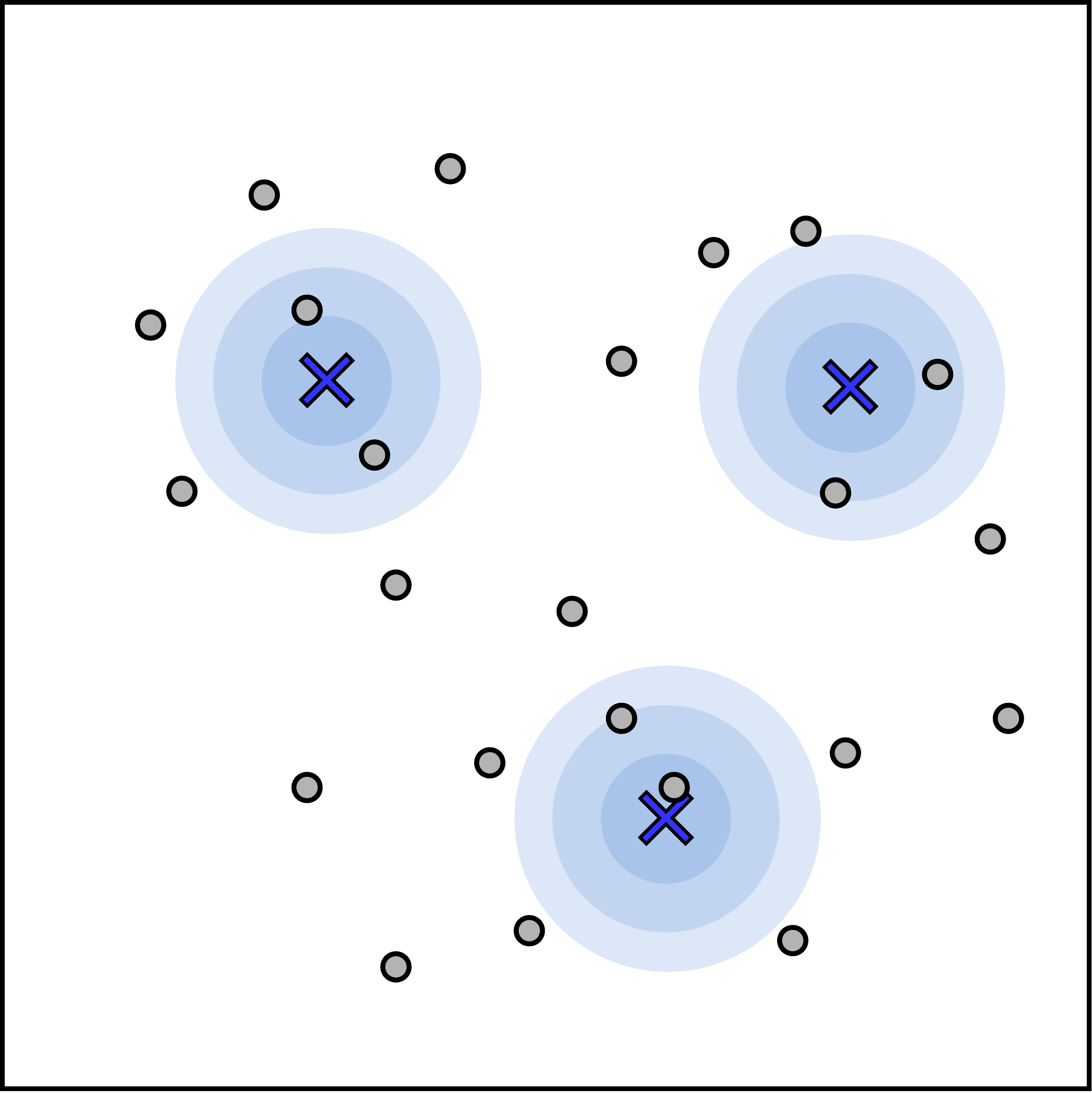}} \hspace*{0.4cm}
\subfigure[Parent $p_2$]{\label{fig:cr02}\includegraphics[width=0.33\textwidth]{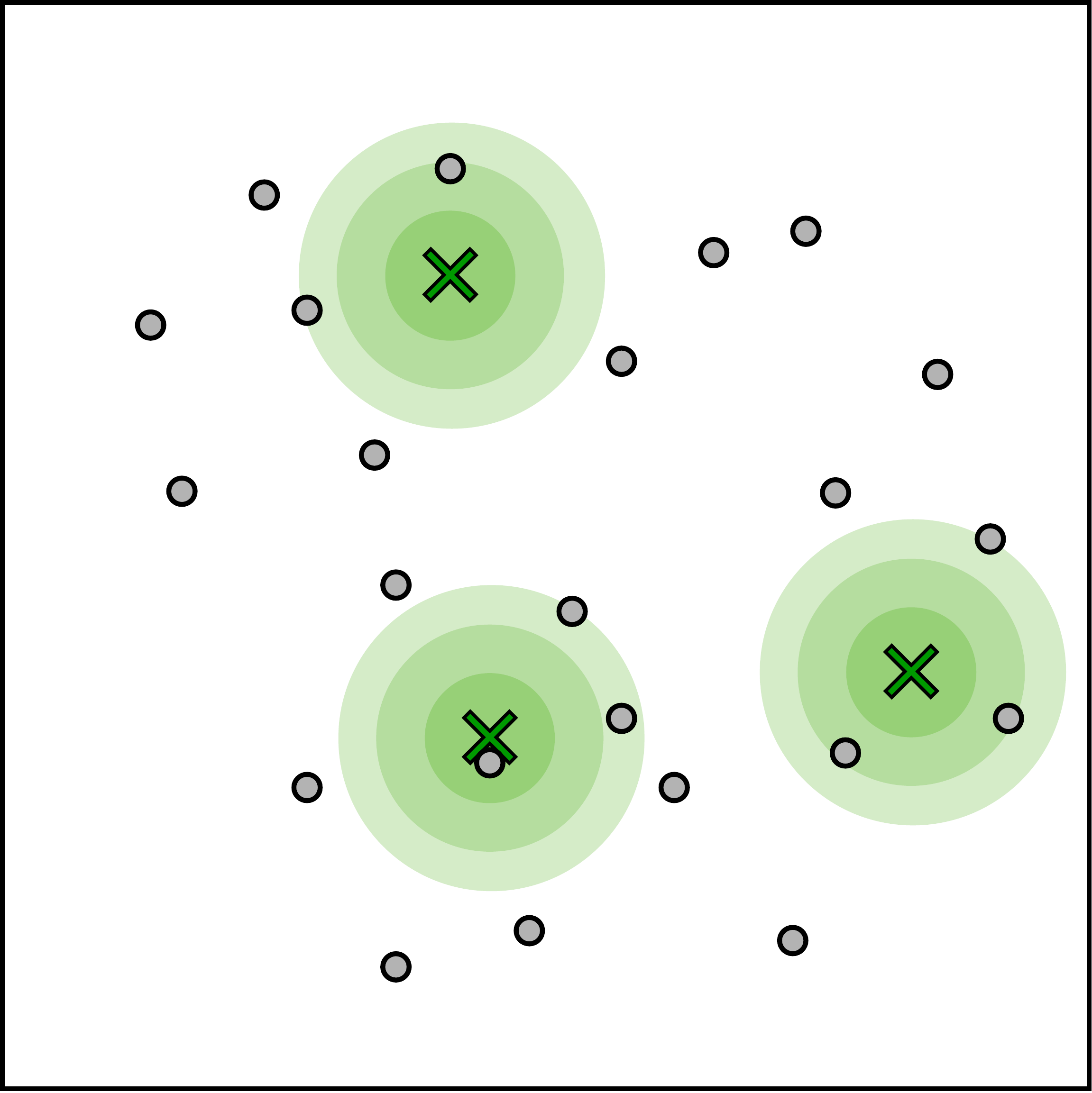}}
\subfigure[Assignment and random selection]{\label{fig:cr03}\includegraphics[width=0.33\textwidth]{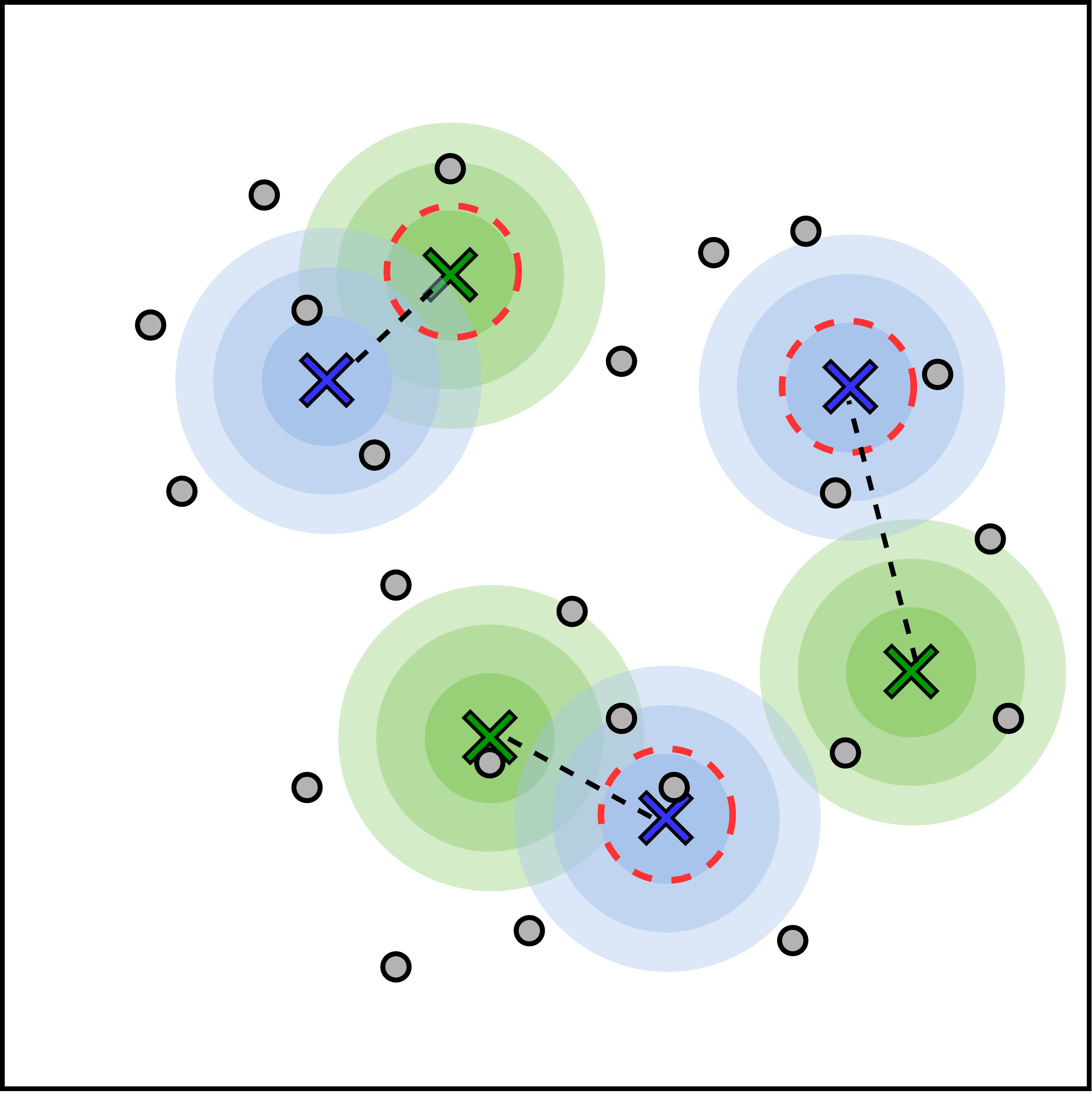}}  \hspace*{0.4cm}
\subfigure[New solution]{\label{fig:cr04}\includegraphics[width=0.33\textwidth]{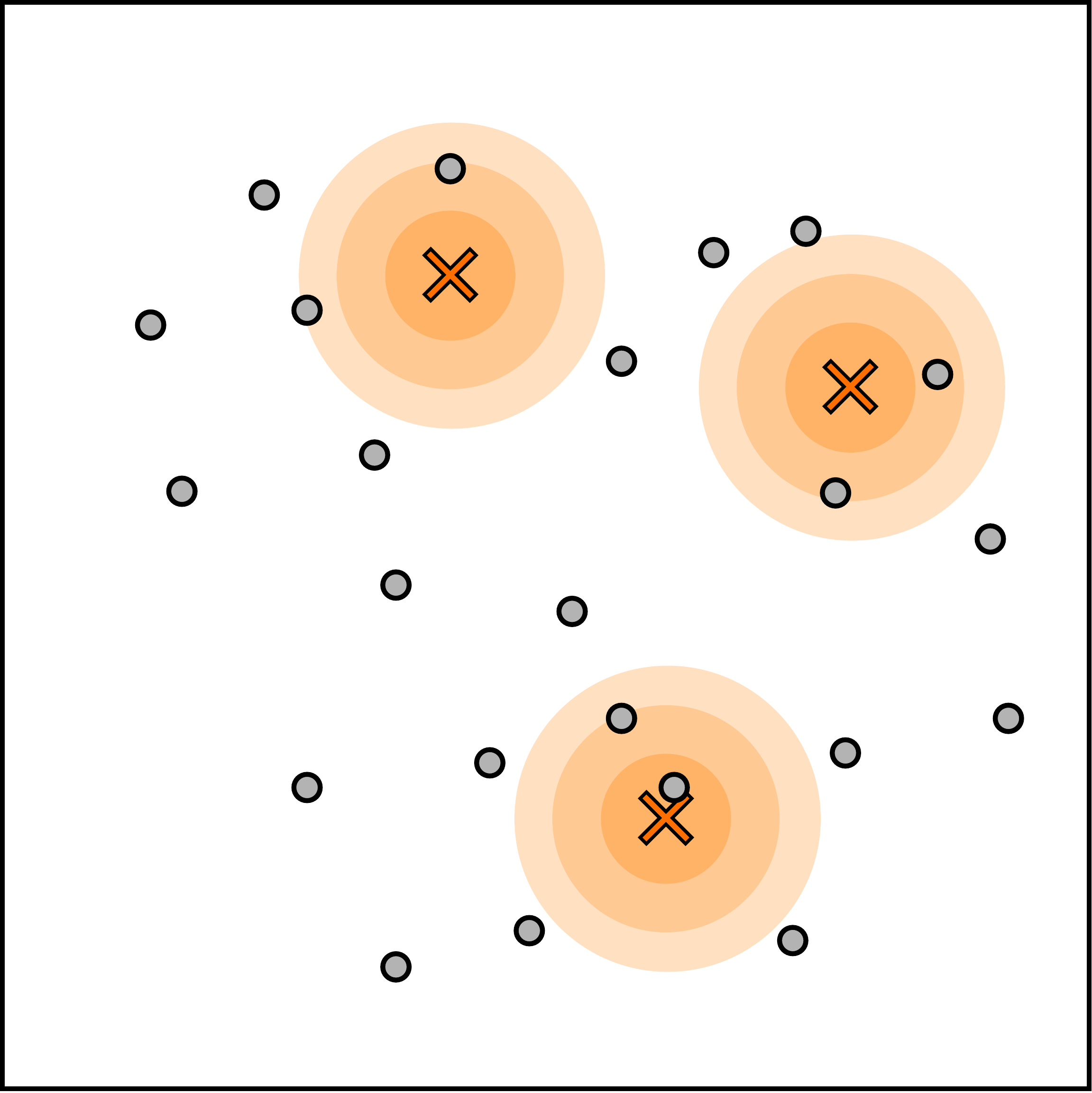}}
\caption{Crossover based on centroid matching: (a) First parent; (b) Second parent; (c) Assignment of centroids and random selection; (d) Resulting offspring.}
\label{fig:crossover}
\end{figure}

The MX is illustrated in Figure \ref{fig:crossover} on the same example as before. This method can be viewed as an extension of the third crossover of \cite{Franti1997}, using an exact matching algorithm instead of a greedy heuristic. The MX has several important properties. First, each center in $\psi_C$ belongs to at least one parent, therefore promoting the transmission of good building blocks \citep{Holland1975}. Second, although any MSSC solution admits $m!$ symmetrical representations, obtained by reindexing its clusters or reordering its centers, the coordinate chromosome generated by the MX will contain the same centers, regardless of this order. In combination with population management mechanisms (Section \ref{sec:population-management}), this helps to avoid the propagation of similar solutions in the population and prevents premature convergence due to a loss of diversity.

\subsection{Mutation}
\label{subsec:mutation}

The MX is deterministic and strictly inherits existing solution features from both parents. In contrast, our mutation operator aims to introduce new randomized solution characteristics. It receives as input the coordinate chromosome $\psi_C$ of the offspring and its mutation parameter $\alpha_C$. It has two steps:
\begin{itemize}[nosep,leftmargin=*]
\item \textsc{Step 1)} It ``mutates'' the mutation parameter as follows:
\begin{equation}
\alpha_{C'} = \max \{ 0, \min \{ \alpha_C + X, 1 \} \},
\end{equation}
where $X$ is a random number selected with uniform probability in $[-0.2,0.2]$. Such a mechanism is typical in evolution strategies and allows an efficient parameter adaptation during the search.

\item \textsc{Step 2)} It uses the newly generated $\alpha_{C'}$ to mutate the coordinate chromosome $\psi_C$ by the biased relocation of a center:
\begin{itemize}[nosep]
	\item Select a random center with uniform probability for removal.
	
	\item Re-assign each sample $p_i$ to the closest remaining center (but do not modify the positions of the centers). Let $d^\textsc{c}_i$ be the distance of each sample $p_i$ from its closest center. 
	
	\item Select a random sample $p_Y$ and add a new center in its position, leading to the new coordinate chromosome $\psi_{C'}$. The selection of $p_Y$ is done by roulette wheel, based on the following mixture of two probability distributions:
\begin{equation}
\label{relocation-proba}
P(Y = i) =   \left( (1-\alpha_{C'}) \times \frac{1}{n} \right) +  \left( \alpha_{C'} \times \frac{\norm{p_i- d^\textsc{c}_i}}{\sum_{j=1}^{n} \norm{p_j- \smash{d^\textsc{c}_j}}} \right)
\end{equation}
\end{itemize}
\end{itemize}

The value of $\alpha_{C'}$ in Equation (\ref{relocation-proba}) drives the behavior of the mutation operator. In the case where $\alpha_{C'} = 0$, all the samples have an equal chance of selection, independently of their distance from the closest center. When~$\alpha_{C'} = 1$, the selection probability is proportional to the distance, similarly to the initialization phase of \textsc{K-means++} \citep{Arthur2007}.
Considering sample-center distances increases the odds of selecting a good position for the new center. However, this could also drive the center positions toward outliers and reduce the solution diversity. For these reasons, $\alpha_{C'}$ is adapted instead of remaining fixed.

\begin{figure}[htb]
\centering
\includegraphics[scale=0.95]{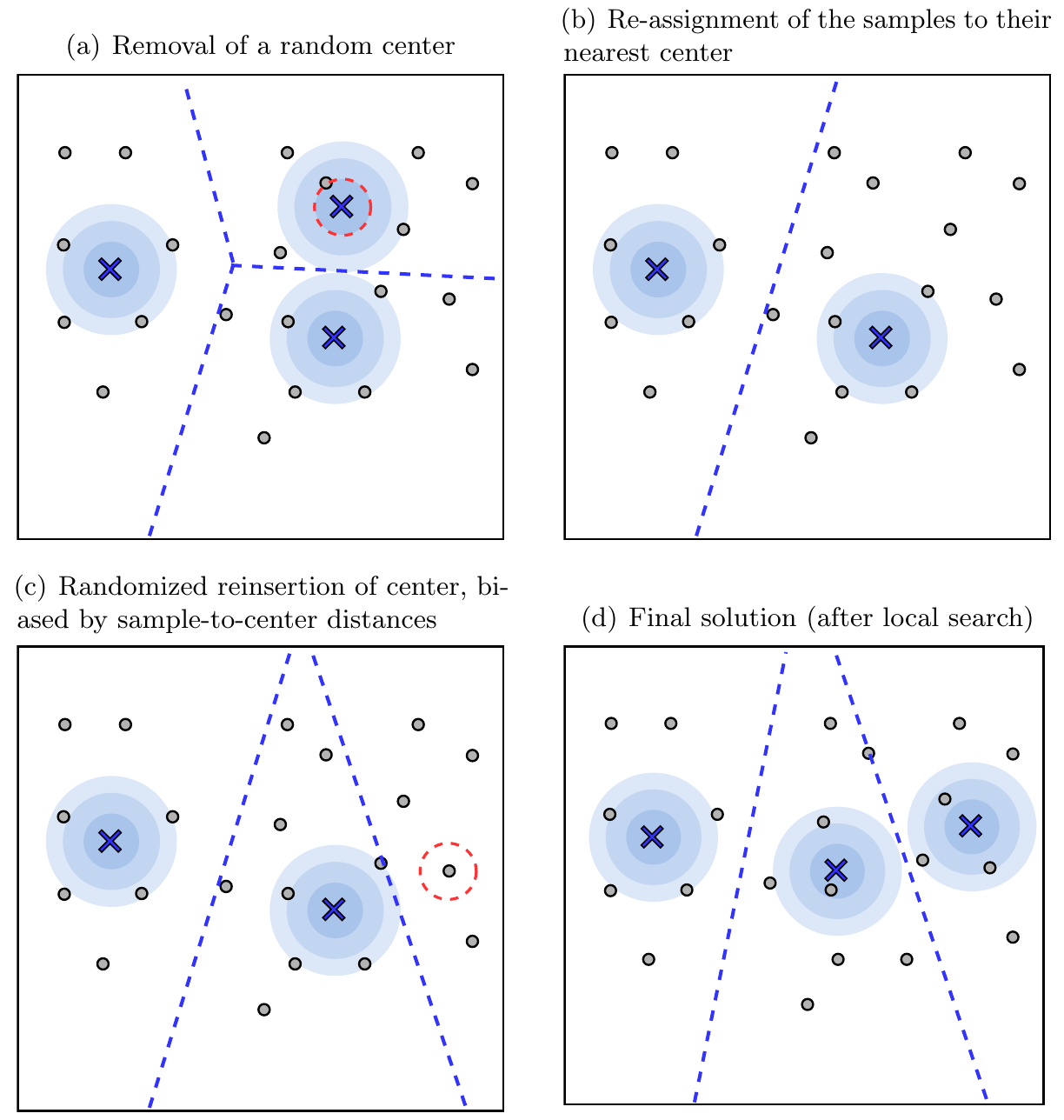}
\caption{Mutation operator by relocation of a single centroid}
\label{fig:mutation}
\end{figure}

The mutation operator requires $\cO(nmd)$ time and is illustrated in Figure \ref{fig:mutation}. The top-right center is selected for removal, and a new center is reinserted on the bottom-right of the figure. These new center coordinates constitute a starting point for the \textsc{K-means} local search discussed in the following section. The first step of \textsc{K-means} will be to reassign each sample to its closest center (Figure \ref{fig:mutation}(c)), which is equivalent to recovering the membership chromosome~$\phi_{C'}$. After a few iterations, \textsc{K-means} converges to a new local minimum (Figure \ref{fig:mutation}(d)), giving a solution that is added to the population.

\subsection{Local Search}
\label{subsec:local-improvement}

Each coordinate chromosome generated through selection, crossover, and mutation serves as a starting point for a local search based on \textsc{K-means}. This algorithm iteratively 1) reassigns each sample to its closest center and 2) moves each center position to the centroid of the samples to which it is assigned. These two steps are iterated until convergence to a local optimum.
For this purpose, we use the fast \textsc{K-means} of \citep{Hamerly2010}. This algorithm has a worst-case complexity of $\mathcal{O}(nmd + m d^2)$ per loop, and it exploits lower bounds on the distances to eliminate many distance evaluations while retaining the same results as the classical \textsc{K-means}.
Moreover, in the exceptional case in which some clusters are left empty after the application of \textsc{K-means} (i.e., one center has no allocated sample), the algorithm selects new center locations using Equation~(\ref{relocation-proba}) and restarts the \textsc{K-means} algorithm to ensure that all the solutions have exactly~$m$~clusters.

\subsection{Population Management}
\label{sec:population-management}

One critical challenge for population-based algorithms is to avoid the premature convergence of the population. Indeed, the elitist selection of parents favors good individuals and tends to reduce the population diversity. This effect is exacerbated by the application of \textsc{K-means} as a local search, since it concentrates the population on a smaller subset of local minima. Once diversity is lost, the chances of generating improving solutions via crossover and mutation are greatly reduced. To overcome this issue, \textsc{HG-means} relies on diversity management operators that find a good balance between elitism and diversification and that allow progress toward unexplored regions of the search space without excluding promising individuals. Similar techniques have been shown to be essential to progress toward high-quality solutions for difficult combinatorial optimization problems \citep{Soerensen2006,Vidal2012}.\\

\noindent
\textbf{Population management.}
The initial population of \textsc{HG-means} is generated by $\Pi_\textsc{max}$ runs of the standard $\textsc{K-means}$ algorithm, in which the initial center locations are randomly selected from the set of samples. Moreover, the mutation parameter of each individual is randomly initialized in $[0,1]$ with an uniform distribution. Subsequently, each new individual is directly included in the population and thus has \emph{some chance} to be selected as a parent by the binary tournament operator, regardless of its quality. Whenever the population reaches a maximum size $\Pi_\textsc{max}$, a \emph{survivor selection} is triggered to retain a subset of $\Pi_\textsc{min}$ individuals.\\

\noindent
\textbf{Survivors selection.} This mechanism selects $\Pi_\textsc{max} - \Pi_\textsc{min}$ individuals in the population for removal. 
To promote population diversity, the removal first focuses on clone individuals. Two individuals $P$ and $P'$ are \emph{clones} if they have the same center positions.
When a pair of clones is detected, one of the two individuals is randomly eliminated.
When no more clones remain, the survivors selection proceeds with the elimination of the worst individuals, until a population of size $\Pi_\textsc{min}$ is obtained.\\

\noindent
\textbf{Complexity analysis.} \textsc{HG-means} has an overall worst-case complexity of $\cO( \Pi_\textsc{max} \Phi_\textsc{Km} +  N_2 \Phi_\textsc{Cross} + N_2  \Phi_\textsc{Mut} + N_2 \Phi_\textsc{Km})$, where $\Phi_\textsc{Cross}$, $\Phi_\textsc{Mut}$, and $\Phi_\textsc{Km}$ represent the time spent in the crossover, mutation, and \textsc{K-means} procedures. The mutation and crossover methods are much faster than the \textsc{K-means} local search in practice, so \textsc{HG-means'} CPU time is proportional to the product of $(\Pi_\textsc{max}+N_2)$ with the time of \textsc{K-means}. Under strict CPU time limits, $\Pi_\textsc{max}$ and $N_2$ could be set to small constants to obtain fast results (see Section \ref{sec:scalability}).
Moreover, since \textsc{HG-means} maintains a population of complete solutions with $m$ clusters, it has good ``anytime behavior'' since it can be interrupted whenever necessary to return the current best solution.

\section{Experimental Analysis}
\label{chap:experiments}

We conducted extensive computational experiments to evaluate the performance of \textsc{HG-means}. After a description of the datasets and a preliminary parameter calibration \mbox{(Sections~\ref{sec:datasets}--\ref{sec:calibration})}, our first analysis focuses on solution quality from the perspective of the MSSC optimization problem (Section \ref{sec:results}). We compare the solution quality obtained by \textsc{HG-means} with that of the current state-of-the-art algorithms, in terms of objective function value and computational time, and we study the sensitivity of the method to changes in the parameters.
Our second analysis concentrates on the scalability of the method, studying how the computational effort grows with the dimensionality of the data and the number of clusters (Section \ref{sec:scalability}).
Finally, our third analysis evaluates the correlation between solution quality from an optimization viewpoint and \myblue{clustering performance via external cluster validity measures}. It compares the performance of \textsc{HG-means}, \textsc{K-means}, and \textsc{K-means++} on a fundamental task, that of recovering the parameters of a non separable mixture of Gaussians (Section~\ref{sec:gaussian}). Yet, we consider feature spaces of medium to high dimensionality (20 to 500) in the presence of a medium to large number of Gaussians (50 to 1,000). We show that the improved solution quality of \textsc{HG-means} directly translates into better \myblue{clustering performance} and more accurate information retrieval.

The algorithms of this paper were implemented in C++ and compiled with G++ 4.8.4. The source code is available at \myblue{\url{https://github.com/danielgribel/hg-means}}.
The experiments were conducted on a single thread of an Intel Xeon X5675 3.07 GHz processor with 8 GB of RAM.

\subsection{Datasets}
\label{sec:datasets}

We evaluated classical and recent state-of-the-art algorithms \citep{Bagirov2008,Bagirov2016,Karmitsa2017,Karmitsa2018,Kivijarvi2003,Ordin2015}, in terms of solution quality for the MSSC objective, on a subset of datasets from the UCI machine learning repository (\url{http://archive.ics.uci.edu/ml/}). 
We collected all the recent datasets used in these studies for a thorough experimental comparison. The resulting 29 datasets, listed in Table \ref{datasets}, arise from a large variety of applications (e.g. cognitive psychology, genomics, and particle physics) and contain numeric features without missing values. 

\begin{table}[htb]
\renewcommand{\arraystretch}{1.05}
\centering
\scalebox{0.85}
{
\begin{tabular}{|c|lrr@{\hspace*{0.7cm}}r@{\hspace*{0.7cm}}c|}
\hline
\textbf{Group} & \textbf{Dataset} & $n$ & $d$ & $n \times d$ & \textbf{Clusters} \\ 
\hline
\multirow{4}{*}{A1} 
& German Towns & 59 & 2 & 118 & \multirow{3}{*}{$m \in \{2, 3, 4...$} \\
& Bavaria Postal 1 &89&3&267& \multirow{3}{*}{$5,6, 7, 8, 9, 10\}$} \\
& Bavaria Postal 2 &89&4&356& \\
& Fisher's Iris Plant &150&4&600& \\
\hline
\multirow{6}{*}{A2} 
& Liver Disorders &345&6&2k& \multirow{5}{*}{$m \in \{2, 5, 10,15...$}\\
& Heart Disease &297&13&4k&  \multirow{5}{*}{$20, 25, 30, 40, 50\}$} \\
& Breast Cancer &683&9&6k& \\
& Pima Indians Diabetes &768&8&6k& \\
& Congressional Voting &435&16&7k& \\
& Ionosphere &351&34&12k& \\
\hline
\multirow{6}{*}{B} 
& TSPLib1060 &1,060&2&2k& \multirow{5}{*}{$m \in \{2, 10, 20, 30...$}\\
& TSPLib3038 &3,038&2&6k& \multirow{5}{*}{$40, 50, 60, 80, 100\}$} \\
& Image Segmentation &2,310&19&44k&\\
& Page Blocks &5,473&10&55k& \\
& Pendigit &10,992&16&176k& \\
& Letters &20,000&16&320k& \\
\hline
\multirow{13}{*}{C} 
& D15112 &15,112&2&30k&\multirow{12}{*}{$m \in \{2, 3, 5,10...$} \\
& Pla85900 &85,900&2&172k&  \multirow{12}{*}{$15, 20, 25\}$}  \\
& EEG Eye State &14,980&14&210k& \\
& Shuttle Control &58,000&9&522k& \\
& Skin Segmentation &245,057&3&735k& \\
& KEGG Metabolic Relation &53,413&20&1M& \\
& 3D Road Network &434,874&3&1M& \\
& Gas Sensor &13,910&128&2M& \\
& Online News Popularity &39,644&58&2M& \\
& Sensorless Drive Diagnosis & 58,509 & 48 &3M& \\
& Isolet & 7,797 & 617 &5M&  \\
& MiniBooNE &130,064&50&7M& \\
& Gisette & 13,500 & 5,000 & 68M & \\
\hline
\end{tabular}
}
\caption{Datasets used for performance comparisons on the MSSC optimization problem}
\label{datasets}
\end{table}

Their dimensions vary significantly, from 59 to 434,874 samples, and from 2~to~5,000 features. Each dataset has been considered with different values of $m$ (number of clusters), leading to a variety of MSSC test set-ups. These datasets are grouped into four classes. Classes A1 and A2 have small datasets with 59 to 768 samples, while Class B has medium datasets with 1,060 to 20,000 samples. These three classes were considered in \cite{Ordin2015}. Class C has larger datasets collected in \cite{Karmitsa2018}, with 13,910 to 434,874 samples and sometimes a large number of features (e.g., \emph{Isolet} and \emph{Gisette}).

\subsection{Parameter Calibration}
\label{sec:calibration}

\textsc{HG-means} is driven by four parameters: two controlling the population size ($\Pi_\textsc{min}$ and~$\Pi_\textsc{max}$) and two controlling the algorithm termination (maximum number of consecutive iterations without improvement $N_1$, and overall maximum number of iterations $N_2$). Changing the termination criteria leads to different nondominated trade-offs between solution quality and CPU time. We therefore set these parameters to consume a smaller CPU time than most existing algorithms on large datasets, while allowing enough iterations to profit from the enhanced exploration abilities of \textsc{HG-means}, leading to $(N_1,N_2) = (500,5000)$. Subsequently, we calibrated the population size to establish a good balance between exploitation (number of generations before termination) and exploration (number of individuals). We compared the algorithm versions with $\Pi_\textsc{min} \in \{5,10,20,40,80\}$ and $\Pi_\textsc{max} \in \{10,20,50,100,200\}$ on medium-sized datasets of class A2 and B with $m \in \{ 2,10,20,30,50 \}$. The setting $(\Pi_\textsc{min},\Pi_\textsc{max})=(10,20)$ had the best performance and forms our baseline configuration. The impact of deviations from this parameter setting will be studied in the next section.

\subsection{Performance on the MSSC Optimization Problem}
\label{sec:results}

We tested \textsc{HG-means} on each MSSC dataset and number of clusters $m$.
Tables \ref{z-avg-all1} and \ref{z-avg-all2} compare its solution quality and CPU time with those of classical and recent state-of-the-art algorithms:
\begin{itemize}[nosep]
\item GKM \citep{Likas2003} -- global K-means;
\item SAGA \citep{Kivijarvi2003} -- self-adaptive genetic algorithm;
\item MGKM \citep{Bagirov2008} -- modified global K-means;
\item MS-MGKM \citep{Ordin2015} -- multi-start modified global K-means;
\item DCClust and MS-DCA \citep{Bagirov2016} -- clustering based on a difference of convex functions;
\item DCD-Bundle \citep{Karmitsa2017} -- diagonal bundle method;
\item LMBM-Clust \citep{Karmitsa2018} -- nonsmooth optimization based on a limited memory bundle method.
\end{itemize}
We also report the results of the classical \textsc{K-means} and \textsc{K-means++} algorithms (efficient implementation of~\cite{Hamerly2010}) for both a single run and the best solution of 5,000 repeated runs with different initial solutions.

The datasets and numbers of clusters $m$ indicated in Table \ref{datasets} lead to 235 test set-ups. For ease of presentation, each line of Tables \ref{z-avg-all1} and \ref{z-avg-all2} is associated with one dataset and displays averaged results over all values of~$m$.
The detailed results of \textsc{HG-means} are available at \url{https://w1.cirrelt.ca/~vidalt/en/research-data.html}.
For each dataset and value of $m$, the solution quality is measured as the percentage gap from the best-known solution (BKS) value reported in all previous articles (from multiple methods, runs, and parameter settings).

\begin{landscape}
\begin{table}[htbp]
\vspace*{-0.75cm}
\hspace*{0cm}
\renewcommand{\arraystretch}{1.1}
{
\setlength\tabcolsep{7pt}
\scalebox{0.75}{
\begin{tabular}{|c|cccc|cccc|cc|cc|cc|cc|cc|}
\hline
&\multicolumn{4}{c|}{\textsc{K-means}}&\multicolumn{4}{c|}{\textsc{K-means++}}&\multicolumn{2}{c|}{GKM}&\multicolumn{2}{c|}{SAGA}&\multicolumn{2}{c|}{MGKM}&\multicolumn{2}{c|}{MS-MGKM}&\multicolumn{2}{c|}{\textsc{HG-means}}\\
&\multicolumn{2}{c}{Single Run}&\multicolumn{2}{c|}{5000 Runs}&\multicolumn{2}{c}{Single Run}&\multicolumn{2}{c|}{5000 Runs}&&&&&&&&&&\\
&Gap&T(s)&Gap&T(s)&Gap&T(s)&Gap&T(s)&Gap&T(s)&Gap&T(s)&Gap&T(s)&Gap&T(s)&Gap&T(s)\\
\hline
German&20.34&0.00&0.00&0.15&15.12&0.00&\textbf{-0.08}&0.14&0.76&0.00&\textbf{-0.08}&0.22&1.00&0.00&0.47&0.01&\textbf{-0.08}&0.02\\
Bavaria1&789.66&0.00&8.28&0.36&10.34&0.00&\textbf{0.00}&0.19&1.03&0.00&\textbf{0.00}&0.26&0.17&0.01&0.07&0.00&\textbf{0.00}&0.02\\
Bavaria2&738.96&0.00&5.63&0.49&37.14&0.00&\textbf{-0.05}&0.25&1.37&0.00&\textbf{-0.05}&0.29&1.37&0.01&0.14&0.00&\textbf{-0.05}&0.03\\
Iris&20.47&0.00&\textbf{0.00}&0.48&10.92&0.00&\textbf{0.00}&0.49&1.48&0.01&\textbf{0.00}&0.53&1.48&0.01&0.09&0.03&\textbf{0.00}&0.09\\
Liver&26.34&0.00&6.90&11.54&8.85&0.00&1.22&9.00&13.44&0.08&-0.06&5.92&12.15&0.21&0.00&1.59&\textbf{-0.94}&1.82\\
Heart&15.24&0.00&3.48&11.08&7.39&0.00&1.52&11.12&1.64&0.07&0.55&16.66&1.63&0.24&0.04&1.51&\textbf{-0.55}&2.17\\
Breast&20.31&0.01&4.95&27.63&5.43&0.00&1.61&22.22&3.29&0.24&1.15&19.21&1.30&0.52&0.00&2.02&\textbf{-0.45}&5.56\\
Pima&21.75&0.01&2.60&35.56&6.12&0.01&0.82&24.72&1.01&0.32&0.87&18.58&0.95&0.64&0.00&3.70&\textbf{-0.11}&5.60\\
Congressional&7.53&0.00&2.34&20.93&5.89&0.01&1.64&23.71&2.61&0.14&0.74&16.34&0.93&0.42&0.00&2.77&\textbf{-0.88}&3.91\\
Ionosphere&14.60&0.01&5.40&30.77&15.56&0.01&3.24&27.97&4.96&0.11&1.05&21.08&0.47&1.43&0.13&1.70&\textbf{-1.59}&5.54\\
TSPLib1060&16.83&0.00&4.04&20.98&11.03&0.00&2.35&18.20&2.63&1.06&0.81&30.83&2.33&1.07&0.20&6.53&\textbf{-0.15}&4.15\\
TSPLib3038&5.54&0.02&1.02&89.16&4.54&0.02&0.74&80.96&1.43&25.32&0.00&35.77&1.07&8.16&0.23&46.31&\textbf{-0.24}&16.66\\
Image&41.67&0.08&16.45&369.38&11.01&0.06&1.63&259.30&1.26&11.67&1.54&143.90&1.31&24.75&0.24&35.34&\textbf{-0.01}&57.51\\
Page&2960.65&0.80&911.35&4538.52&14.13&0.08&1.11&340.92&1.25&129.99&61.58&134.88&0.93&97.26&0.07&31.72&\textbf{-0.96}&143.67\\
Pendigit&3.31&0.50&0.36&2504.67&2.50&0.50&0.22&2275.14&0.24&263.04&0.74&607.68&0.16&434.83&0.04&352.36&\textbf{-0.18}&461.13\\
Letters&1.98&1.26&0.14&6209.46&1.35&1.35&0.10&6751.95&0.35&1102.38&0.42&1114.36&0.13&1859.64&0.01&908.70&\textbf{-0.18}&1326.05\\
\hline
Avg. Gap&\multicolumn{2}{c}{294.07}&\multicolumn{2}{c|}{60.81}&\multicolumn{2}{c}{10.46}&\multicolumn{2}{c|}{1.00}&\multicolumn{2}{c|}{2.42}&\multicolumn{2}{c|}{4.33}&\multicolumn{2}{c|}{1.71}&\multicolumn{2}{c|}{0.11}&\multicolumn{2}{c|}{\textbf{-0.40}}\\
CPU&\multicolumn{4}{c|}{Xe 3.07 GHz}&\multicolumn{4}{c|}{Xe 3.07 GHz}&\multicolumn{2}{c|}{Core2 2.5 GHz}&\multicolumn{2}{c|}{Xe 3.07 GHz}&\multicolumn{2}{c|}{Core2 2.5 GHz}&\multicolumn{2}{c|}{Core2 2.5 GHz}&\multicolumn{2}{c|}{Xe 3.07 GHz}\\
Passmark&\multicolumn{4}{c|}{1403 (1.00)}&\multicolumn{4}{c|}{1403 (1.00)}&\multicolumn{2}{c|}{976 (0.70)}&\multicolumn{2}{c|}{1403 (1.00)}&\multicolumn{2}{c|}{976 (0.70)}&\multicolumn{2}{c|}{976 (0.70)}&\multicolumn{2}{c|}{1403 (1.00)}\\
\hline
\end{tabular}
}}
\caption{Performance comparison for small and medium MSSC datasets}\label{z-avg-all1}
\vspace{1cm}
\hspace*{-1.3cm}
\setlength\tabcolsep{3.0pt}
\scalebox{0.72}{
\begin{tabular}{|c|cccc|cccc|cc|cc|cc|cc|cc|cc|cc|cc|}
\hline
 & \multicolumn{4}{c|}{{\textsc{K-means}}} & \multicolumn{4}{c|}{\textsc{K-means++}} &\multicolumn{2}{c|}{GKM} & \multicolumn{2}{c|}{\textsc{SAGA}} & \multicolumn{2}{c|}{\textsc{LMBM}} & \multicolumn{2}{c|}{MS-MGKM} & \multicolumn{2}{c|}{DCClust} & \multicolumn{2}{c|}{MS-DCA} & \multicolumn{2}{c|}{DCD-Bundle} & \multicolumn{2}{c|}{\textsc{HG-means}} \\
 & \multicolumn{2}{c}{Single Run} & \multicolumn{2}{c|}{5000 Runs} & \multicolumn{2}{c}{Single Run} & \multicolumn{2}{c|}{5000 Runs} & & & & & & & & & & & & & & & & \\
 & Gap & T(s) & Gap & T(s) & Gap & T(s) & Gap & T(s) & Gap & T(s) & Gap & T(s) & Gap & T(s) & Gap & T(s) & Gap & T(s) & Gap & T(s) & Gap & T(s) & Gap & T(s)\\
\hline
D15112&1.60&0.04&\textbf{0.00}&160.69&1.18&0.03&\textbf{0.00}&146.53&0.34&43.25&0.19&47.29&0.34&4.58&0.13&11.28&0.12&16.26&0.13&35.73&0.13&9.60&\textbf{0.00}&17.52\\
Pla85900&0.46&0.31&\textbf{-0.02}&1115.56&0.79&0.26&\textbf{-0.02}&1060.07&0.25&2023.84&0.15&260.18&0.95&22.36&0.10&2094.58&0.14&200.30&0.09&1416.24&0.15&185.61&\textbf{-0.02}&198.14\\
Eye&880402.99&0.39&0.00&1522.87&48.95&0.23&0.58&961.63&0.81&161.18&0.04&212.00&0.75&6.62&0.74&17.59&0.89&41.23&0.75&121.77&0.98&19.45&\textbf{-0.02}&196.43\\
Shuttle&181.15&0.81&121.42&2832.02&22.49&0.25&-0.66&911.64&0.35&1954.72&-0.90&475.43&0.10&4.55&0.41&89.27&0.46&227.47&0.41&4722.90&1.71&312.10&\textbf{-0.91}&97.84\\
Skin&9.63&0.60&0.05&2304.23&7.71&0.43&-0.38&1728.25&0.28&22518.92&0.13&844.14&3.93&14.41&0.63&3021.33&0.32&1233.00&0.33&8774.03&0.32&1259.01&\textbf{-0.41}&230.25\\
Kegg&94.45&4.21&76.24&17988.47&6.96&0.58&-0.49&1204.77&1.85&4147.46&3.54&686.18&1.52&10.78&1.52&445.03&1.18&488.31&1.02&3442.98&0.89&576.76&\textbf{-0.51}&244.37\\
3Droad&0.23&5.55&\textbf{0.00}&35237.41&0.28&2.99&\textbf{0.00}&13437.58&\textbf{0.00}$^\dagger$&42431.72$^\dagger$&0.02&1233.00&\textbf{0.00}&63.21&0.54&60180.29&0.56&4924.72&0.44&31325.93&0.01&4872.71&\textbf{0.00}&2862.09\\
Gas&21.42&1.57&1.87&5182.01&7.27&1.12&-0.17&2482.24&0.24&1550.29&-0.20&983.33&0.86&86.16&0.02&256.74&0.24&814.34&0.05&2404.14&0.55&627.13&\textbf{-0.22}&521.57\\
Online&18.65&1.57&0.51&5245.66&19.74&1.33&\textbf{-0.17}&2653.76&--&--&-0.15&1160.02&4.54&96.18&0.12&795.22&0.00&1509.81&--&--&0.26&1600.50&\textbf{-0.17}&473.23\\
Sensorless&155.50&4.91&46.02&18798.70&19.06&2.50&\textbf{-0.44}&8237.59&--&--&1.16&2004.10&1.18&25.24&0.27&1249.96&2.58&2133.54&--&--&--&--&-0.41&1077.67\\
Isolet&1.93&3.61&\textbf{-0.21}&11082.28&1.33&3.74&\textbf{-0.21}&12587.68&--&--&\textbf{-0.21}&3961.39&0.49&97.59&0.39&677.14&0.32&1672.82&--&--&--&--&\textbf{-0.21}&1846.70\\
Miniboone&40992.86&15.63&-0.07&57148.73&1.75&9.63&\textbf{-0.10}&27611.08&--&--&-0.07&4883.13&3.50&88.41&0.29&7559.34&0.24&9656.14&--&--&0.23&9291.40&\textbf{-0.10}&2941.52\\
Gisette&-0.47&77.17&--&--&-0.47&96.63&--&--&--&--&--&--&0.03&1871.67&0.01$^\ddagger$&39504.13$^\ddagger$&0.00$^\dagger$&49847.13$^\dagger$&--&--&--&--&\textbf{-0.52}&22279.47\\
\hline
Avg. Gap$^*$ & \multicolumn{2}{c}{110088.99} &  \multicolumn{2}{c|}{24.94} &  \multicolumn{2}{c}{11.96} &  \multicolumn{2}{c|}{-0.14}  &  \multicolumn{2}{c|}{0.52} &  \multicolumn{2}{c|}{0.37} & \multicolumn{2}{c|}{1.05} & \multicolumn{2}{c|}{0.51} &  \multicolumn{2}{c|}{0.49} &  \multicolumn{2}{c|}{0.40} &  \multicolumn{2}{c|}{0.59} & \multicolumn{2}{c|}{\textbf{-0.26}} \\

CPU & \multicolumn{4}{c|}{Xe 3.07 GHz} & \multicolumn{4}{c|}{Xe 3.07 GHz} & \multicolumn{2}{c|}{I5 2.9 GHz}& \multicolumn{2}{c|}{Xe 3.07 GHz} & \multicolumn{2}{c|}{I7 4.0 GHz} & \multicolumn{2}{c|}{I7 4.0 GHz} & \multicolumn{2}{c|}{I7 4.0 GHz} & \multicolumn{2}{c|}{I5 2.9 GHz} & \multicolumn{2}{c|}{I5 1.6/2.7 GHz} & \multicolumn{2}{c|}{Xe 3.07 GHz}\\

Passmark & \multicolumn{4}{c|}{1403 (1.00)} & \multicolumn{4}{c|}{1403 (1.00)} &  \multicolumn{2}{c|}{1859 (1.32)} & \multicolumn{2}{c|}{1403 (1.00)} & \multicolumn{2}{c|}{2352 (1.68)} & \multicolumn{2}{c|}{2352 (1.68)} & \multicolumn{2}{c|}{2352 (1.68)} & \multicolumn{2}{c|}{1859 (1.32)} & \multicolumn{2}{c|}{1432 (1.02)} & \multicolumn{2}{c|}{1403 (1.00)}\\ \hline

\multicolumn{21}{l}{
\begin{small}
\hspace*{-0.5cm} * Considering the subset of 8 instances which is common to all methods
\end{small}
\vspace*{-0.15cm}}\\

\multicolumn{21}{l}{
\begin{small}
\hspace*{-0.5cm} $^\dagger$ Considering $m \in \{2, 3, 5\}$
\end{small}
\vspace*{-0.15cm}}\\

\multicolumn{21}{l}{
\begin{small}
\hspace*{-0.5cm} $^\ddagger$ Considering $m \in \{2, 3, 5, 10\}$
\end{small}
\vspace*{-1.25cm}}\\

\end{tabular}
}
\caption{Performance comparison for large MSSC datasets}
\label{z-avg-all2}
\end{table}
\end{landscape}

This gap is expressed as \mbox{Gap($\%$) $= 100 \times (z - z_\textsc{bks})/z_\textsc{bks}$,} where~$z$ represents the solution value of the method considered, and $z_\textsc{bks}$ is the BKS value.
A negative gap means that the solutions for this dataset are better than the best solutions found previously.
Finally, the last two lines indicate the CPU model used in each study, along with the time-scaling factor (based on the Passmark benchmark) representing the ratio between its speed and that of our processor. All time values in this article have been multiplied by these scaling factors to account for processor differences.

\textsc{HG-means} produces solutions of remarkable quality, with an average gap of $-0.40\%$ and $-0.26\%$ on the small and large datasets, respectively.
This means that its solutions are better, on average, than the best solutions ever found.
For all datasets, \textsc{HG-means} achieved the best gap value.
The statistical significance of these improvements is confirmed by pairwise Wilcoxon tests between the results of \textsc{HG-means} and those of other methods (with p-values $< 10^{-8}$). Over all 235 test set-ups (dataset $\times$ number of cluster combinations), \textsc{HG-means} found 113 solutions better than the BKS, 116 solutions of equal quality, and only five solutions of lower quality.
We observe that the improvements are also correlated with the size of the datasets. For the smallest ones, all methods perform relatively well. However, for more complex applications involving a larger number of samples, a feature space of higher dimension, and more clusters, striking differences in solution quality can be observed between the state-of-the-art methods.

These experiments also confirm the known fact that a single run of \textsc{K-means} or \textsc{K-means++} does not usually find a good local minimum of the MSSC problem, as shown by gap values that can become arbitrarily high. For the Eye and Miniboone datasets, in particular, a misplaced center can have large consequences in terms of objective value. The probability of falling into such a case is high in a single run, but it can be reduced by performing repeated runs and retaining the best solution. Nevertheless, even 5,000 independent runs of \textsc{K-means} or \textsc{K-means++} are insufficient to achieve solutions of a quality comparable to that of \textsc{HG-means}.

In terms of computational effort, \textsc{HG-means} is generally faster than SAGA, MS-MGKM, DCClust, MS-DCA, and DCD-Bundle (the current best methods in terms of solution quality), but slower than LMBM-Clust, since this method is specifically designed and calibrated to return quick solutions.
It is also faster than a repeated \textsc{K-means} or \textsc{K-means++} algorithm with 5000 restarts, i.e., a number of restarts equal to the maximum number of iterations of the algorithm. This can be partly explained by the fact that the solutions generated by the exact matching crossover require less time to converge via \textsc{K-means} than initial sample points selected according to some probability distributions. Moreover, a careful exploitation of the search history, represented by the genetic material of high-quality parent solutions, makes the method more efficient and accurate.

Finally, we measured the sensitivity of \textsc{HG-means} to changes in its parameters: $(\Pi_\textsc{min},\Pi_\textsc{max})$ defining the population-size limits, and $(N_1,N_2)$ defining the termination criterion.
In Table~\ref{table:sensitivity1}, we fix the termination criterion to $(N_1,N_2) = (500,5000)$ and consider a range of population-size parameters, reporting the average gap and median time over all datasets for each configuration.
The choice of $\Pi_\textsc{min}$ and $\Pi_\textsc{max}$ appears to have only a limited impact on solution quality and CPU time: regardless of the parameter setting, \textsc{HG-means} returns better average solutions than all individual best known solutions collected from the literature. Some differences can still be noted between configurations: as highlighted by pairwise Wilcoxon tests, every configuration underlined in the table performs better than every non-underlined one (with p-values $\leq$ 0.018). Letting the population rise to the double of the minimum population size ($\Pi_\textsc{max} \approx 2 \times \Pi_\textsc{min}$) before survivors selection is generally a good policy. Moreover, we observe that smaller populations trigger a faster convergence but at the risk of reducing diversity, whereas excessively large populations (i.e., $\Pi_\textsc{max} = 200$) unnecessarily spread the search effort, with an adverse impact on solution quality.

\begin{table}[htb]
\centering
\renewcommand{\arraystretch}{1.3}
\setlength\tabcolsep{6pt}
\scalebox{0.8}{
\begin{tabular}{|r|r@{\hspace*{0.1cm}}rrrrr|rrrrr|}
\hline
&\multicolumn{6}{c|}{\textbf{Average Gap\,(\%)}}&\multicolumn{5}{c|}{\textbf{Median Time\,(s)}}\\
&$\Pi_\textsc{max} $&= 10&20&50&100&200&10&20&50&100&200\\
\hline
$\Pi_\textsc{min}=5$&&\underline{-0.31}&\underline{-0.31}&-0.28&-0.24&-0.20&12.03&11.81&13.54&14.73&17.79\\
10&&&\underline{-0.32}&-0.28&-0.26&-0.23&&12.03&11.53&14.29&16.76\\
20&&&&\underline{-0.31}&-0.25&-0.22&&&15.27&15.94&13.44\\
40&&&&&-0.25&-0.14&&&&13.35&15.19\\
80&&&&&&-0.11&&&&&19.48\\
\hline
\end{tabular}
}
\caption{Sensitivity of \textsc{HG-means} to changes of population-size parameters}
\label{table:sensitivity1}
\end{table}

\begin{table}[htb]
\centering
\renewcommand{\arraystretch}{1.3}
\setlength\tabcolsep{7pt}
\scalebox{0.8}{
\begin{tabular}{|r|rrrrr|rrrrr|}
\hline
&\multicolumn{5}{c|}{\textbf{Average Gap\,(\%)}}&\multicolumn{5}{c|}{\textbf{Median Time\,(s)}}\\
$N_2 =10 \times N_1$&$N_1$ = 50&100&250&500&1000&50&100&250&500&1000\\
\hline
$(\Pi_\textsc{min},\Pi_\textsc{max})$  = (5,10)&0.20&-0.07&-0.25&-0.31&-0.35&1.87&3.31&7.40&12.03&29.10\\
(10,20)&0.52&-0.02&-0.22&-0.32&-0.36&1.54&4.13&8.25&12.03&29.90\\
\hline
\end{tabular}
}
\caption{Sensitivity of \textsc{HG-means} to changes of the termination criterion}
\label{table:sensitivity2}
\end{table}

In Table~\ref{table:sensitivity2}, we retain two of the best population-size configurations $(\Pi_\textsc{min},\Pi_\textsc{max}) \in \{(5,10),(10,20)\}$ and vary the termination criterion. Naturally, the quality of the solutions improves with longer runs, but even a short termination criterion such as $(N_1,N_2) = (100,1000)$ already gives good solutions, with an average gap of $-0.07\%$. Finally, reducing the population size to $(\Pi_\textsc{min},\Pi_\textsc{max}) = (5,10)$ for short runs allows us to better exploit a limited number of iterations, whereas the baseline setting performs slightly better for longer runs.

\subsection{Scalability}
\label{sec:scalability}

The solution quality and computational efficiency of most clustering algorithms is known to deteriorate as the number of clusters $m$ grows, since this leads to more complex combinatorial problems with numerous local minima. To evaluate how \textsc{HG-means} behaves in these circumstances, we conduct additional experiments focused on the large datasets of class C.
Table~\ref{table:detailed} reports the solution quality and CPU time of \textsc{HG-means} for each dataset as a function of $m$.
Moreover, to explore the case where the CPU time is more restricted, Table~\ref{table:detailed:fast} reports the same information for a \emph{fast} \textsc{HG-means} configuration where $(\Pi_\textsc{min},\Pi_\textsc{max}) = (5,10)$ and $(N_1,N_2) = (50,500)$.

\begin{table}[htb]
\renewcommand{\arraystretch}{1.15}
\setlength\tabcolsep{5pt}
\vspace*{0.2cm}
\centering
\scalebox{0.75}{
\begin{tabular}{|l|ccccccc|@{\hspace*{0cm}}H@{\hspace*{0.15cm}}ccccccc|}
\hline
&\multicolumn{7}{c|}{\textbf{Gap\,(\%)}}&&\multicolumn{7}{c|}{\textbf{Time\,(s)}}\\
 &$m$ = 2&3&5&10&15&20&25&&2&3&5&10&15&20&25\\
\hline
D15112 &0.00&0.00&0.00&0.00&0.00&0.00&-0.03&&2.80&5.23&4.97&9.36&37.12&22.45&40.74\\
Pla85900 &0.00&0.00&0.00&0.00&0.00&-0.02&-0.13&&23.94&34.96&95.62&135.77&131.92&349.98&614.79\\
Eye&0.00&0.00&0.00&-0.01&0.00&0.00&-0.16&&4.38&5.22&13.60&91.01&121.20&509.55&630.08\\
Shuttle &0.00&0.00&0.00&-0.02&0.00&-3.67&-2.68&&17.53&18.93&22.91&45.74&63.10&175.70&340.98\\
Skin &0.00&0.00&0.00&0.00&-1.63&-0.89&-0.38&&66.23&90.92&96.21&176.57&336.70&213.78&631.37\\
Kegg &0.00&0.00&0.00&0.00&-1.29&-1.26&-1.03&&41.45&66.68&90.63&117.53&226.37&424.89&743.02\\
3Droad &0.00&0.00&0.00&0.00&0.00&0.00&0.00&&444.22&535.59&498.42&2824.24&2582.08&5888.17&7261.91\\
Gas &0.00&0.00&0.00&-0.18&-0.94&-0.21&-0.18&&93.20&87.71&156.56&222.77&827.35&920.49&1342.93\\
Online &0.00&0.00&0.00&0.00&0.00&-0.01&-1.20&&109.80&87.27&190.75&333.71&285.21&1154.91&1150.95\\
Sensorless &0.00&0.00&0.00&-2.42&0.00&-0.63&0.17&&88.43&256.11&236.20&646.31&1004.29&2179.93&3132.41\\
Isolet &0.00&0.00&0.00&0.00&-0.15&-0.39&-0.96&&255.04&322.27&751.86&748.94&1992.24&2521.89&6334.68\\
Miniboone &0.00&0.00&0.00&0.00&0.00&-0.12&-0.57&&209.08&565.88&585.91&1329.57&4758.04&5061.36&8080.78\\
Gisette&0.00&0.00&-0.02&0.00&-0.51&-1.85&-1.28&&2304.41&3896.54&10964.95&20617.57&31767.12&39283.79&47121.92\\
\hline
\end{tabular}}
\caption{Performance of \textsc{HG-means} as a function of the number of clusters}
\label{table:detailed}
\end{table}

\begin{table}[htbp]
\renewcommand{\arraystretch}{1.15}
\setlength\tabcolsep{5.8pt}
\vspace*{0.2cm}
\centering
\scalebox{0.75}{
\begin{tabular}{|l|ccccccc|@{\hspace*{0cm}}H@{\hspace*{0.15cm}}ccccccc|}
\hline
&\multicolumn{7}{c|}{\textbf{Gap\,(\%)}}&&\multicolumn{7}{c|}{\textbf{Time\,(s)}}\\
 &$m$ = 2&3&5&10&15&20&25&&2&3&5&10&15&20&25\\
\hline
D15112 &0.00&0.00&0.00&0.00&0.00&0.00&-0.03&&0.29&0.95&0.54&1.77&1.93&4.53&13.84\\
Pla85900 &0.00&0.00&0.00&0.00&0.00&-0.02&-0.13&&3.21&3.86&5.22&11.84&29.58&42.44&53.23\\
Eye &0.00&0.00&0.00&-0.01&0.00&0.00&-0.16&&0.68&1.08&1.10&12.28&43.89&31.88&92.41\\
Shuttle &0.00&0.00&0.00&0.46&0.02&-3.55&-2.68&&2.03&2.36&6.50&12.40&35.60&57.68&67.04\\
Skin &0.00&0.00&0.00&0.00&-1.63&-0.89&-0.19&&5.99&9.24&14.37&14.45&35.76&25.49&96.52\\
Kegg &0.00&0.00&0.00&0.00&-1.29&-1.25&-0.93&&4.05&6.17&5.91&25.21&42.94&115.85&73.93\\
3Droad &0.00&0.00&0.00&0.00&0.00&0.00&0.00&&18.37&30.66&38.11&394.55&166.58&288.66&619.88\\
Gas &0.00&0.00&0.00&-0.18&-0.94&-0.21&-0.18&&4.98&11.07&13.74&69.15&94.68&106.94&182.29\\
Online &0.00&0.00&0.00&0.00&0.00&-0.01&-1.20&&8.61&18.12&23.07&39.39&59.10&62.54&218.82\\
Sensorless &0.00&0.00&0.00&-2.42&0.00&-0.63&0.00&&13.92&16.43&22.92&102.44&217.94&397.28&429.3\\
Isolet &0.00&0.00&0.00&0.00&-0.08&-0.39&-0.79&&29.37&43.21&91.75&256.56&415.56&443.05&199.26\\
Miniboone &0.00&0.00&0.00&0.00&0.00&-0.12&-0.57&&17.46&37.82&39.95&246.58&647.88&731.73&1469.62\\
Gisette &0.00&0.00&-0.02&0.53&-0.16&-1.62&-1.14&&232.87&1067.71&1252.21&3190.44&5671.95&13542.21&12568.59\\
\hline
\end{tabular}}
\caption{Performance of a fast configuration of \textsc{HG-means} as a function of the number of clusters}
\label{table:detailed:fast}
\end{table}

As observed in Table~\ref{table:detailed}, \textsc{HG-means} retrieves or improves the BKS for all datasets and values of $m$.
Significant improvements are more frequently observed for larger values of $m$.
A likely explanation is that the global minimum has already been found for most datasets with a limited number of clusters, whereas previous methods did not succeed in finding the global optimum for larger values of $m$.
In terms of computational effort, there is a visible correlation between the number of clusters $m$ and the CPU time.
Power law regressions of the form $f(m) = \alpha m^\beta$ indicate that the computational effort of \textsc{HG-means} grows as $\Theta(m^{2.09})$ for Eye State, $\Theta(m^{1.38})$ for Miniboone, and $\Theta(m^\beta)$ for $\beta \leq 1.29$ in all other cases.
Similarly, for $m=10$, fitting the CPU time of the method as a power law of the form $g(n,d) = \alpha n^{\beta} d^{\,\gamma}$ indicates that the measured CPU time of \textsc{HG-means} grows as $\cO(n^{1.08} d^{\, 0.88})$, i.e., linearly with the number of samples and the dimension of the feature space.

\begin{figure}[p]
\thispagestyle{empty}
\centering

\vspace*{-0.6cm}

\begin{minipage}{0.322\textwidth}
\scalebox{0.9}{\hspace*{0.55cm}D15112 ($n = 15112$, $d = 2$)}
\vspace*{0.08cm}
\includegraphics[width=\textwidth]{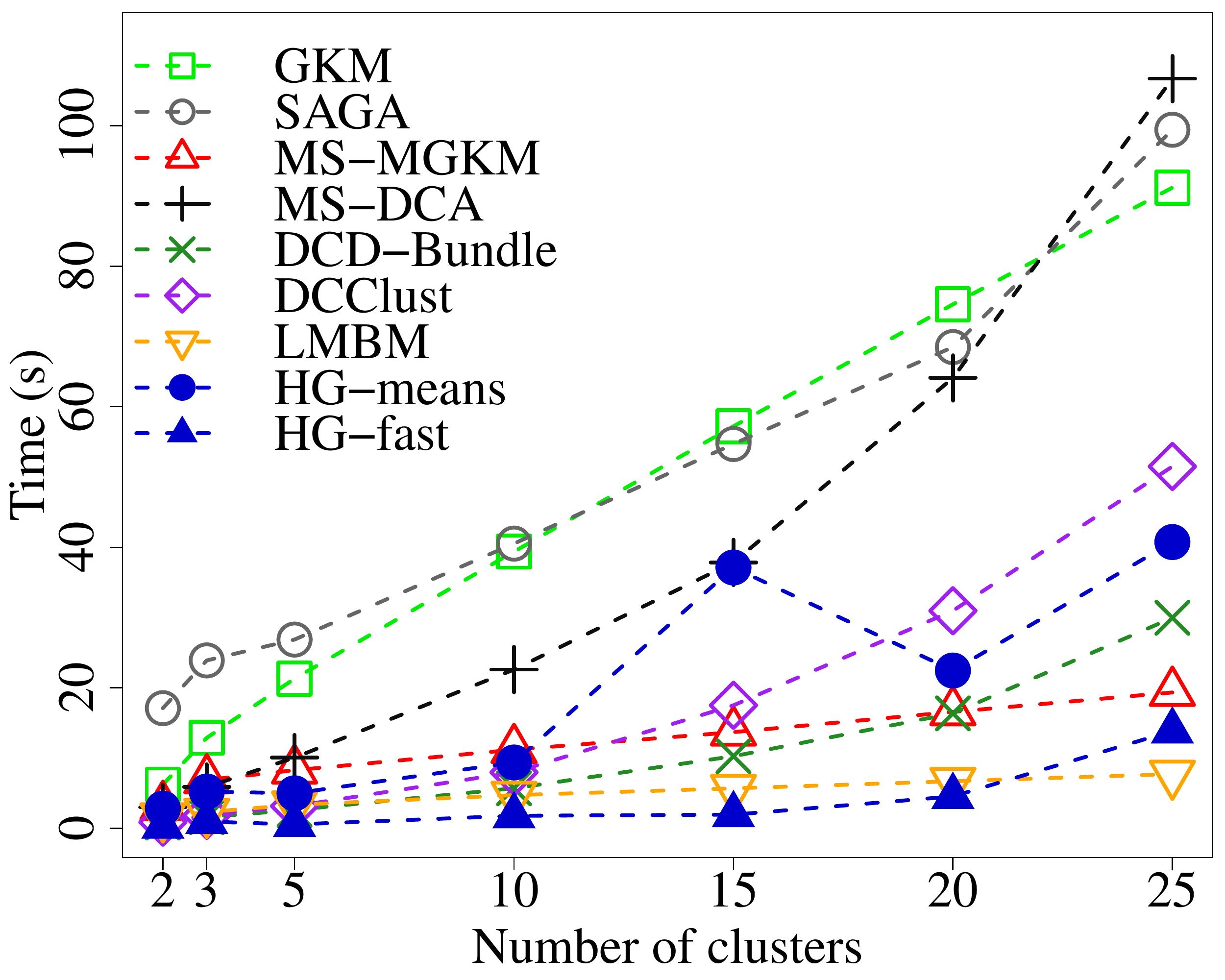}
\end{minipage}
\hspace*{0cm}
\begin{minipage}{0.322\textwidth}
\scalebox{0.9}{\hspace*{0.55cm}Pla85900 ($n = 85900$, $d = 2$)}
\vspace*{0.08cm}
\includegraphics[width=\textwidth]{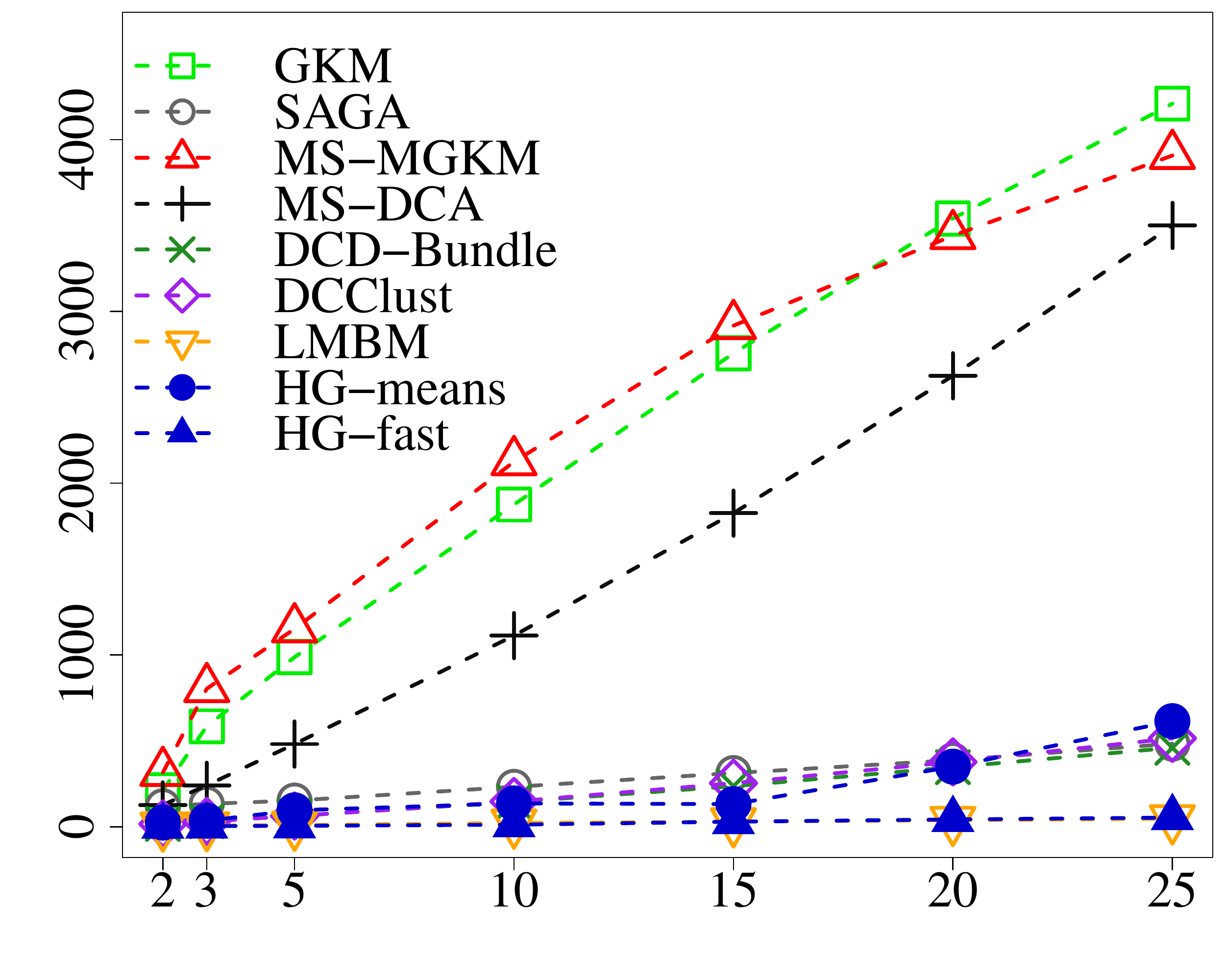}
\end{minipage}
\hspace*{0cm}
\begin{minipage}{0.322\textwidth}
\scalebox{0.9}{\hspace*{0.55cm}EEG Eye ($n = 14980$, $d = 14$)}
\vspace*{0.08cm}
\includegraphics[width=\textwidth]{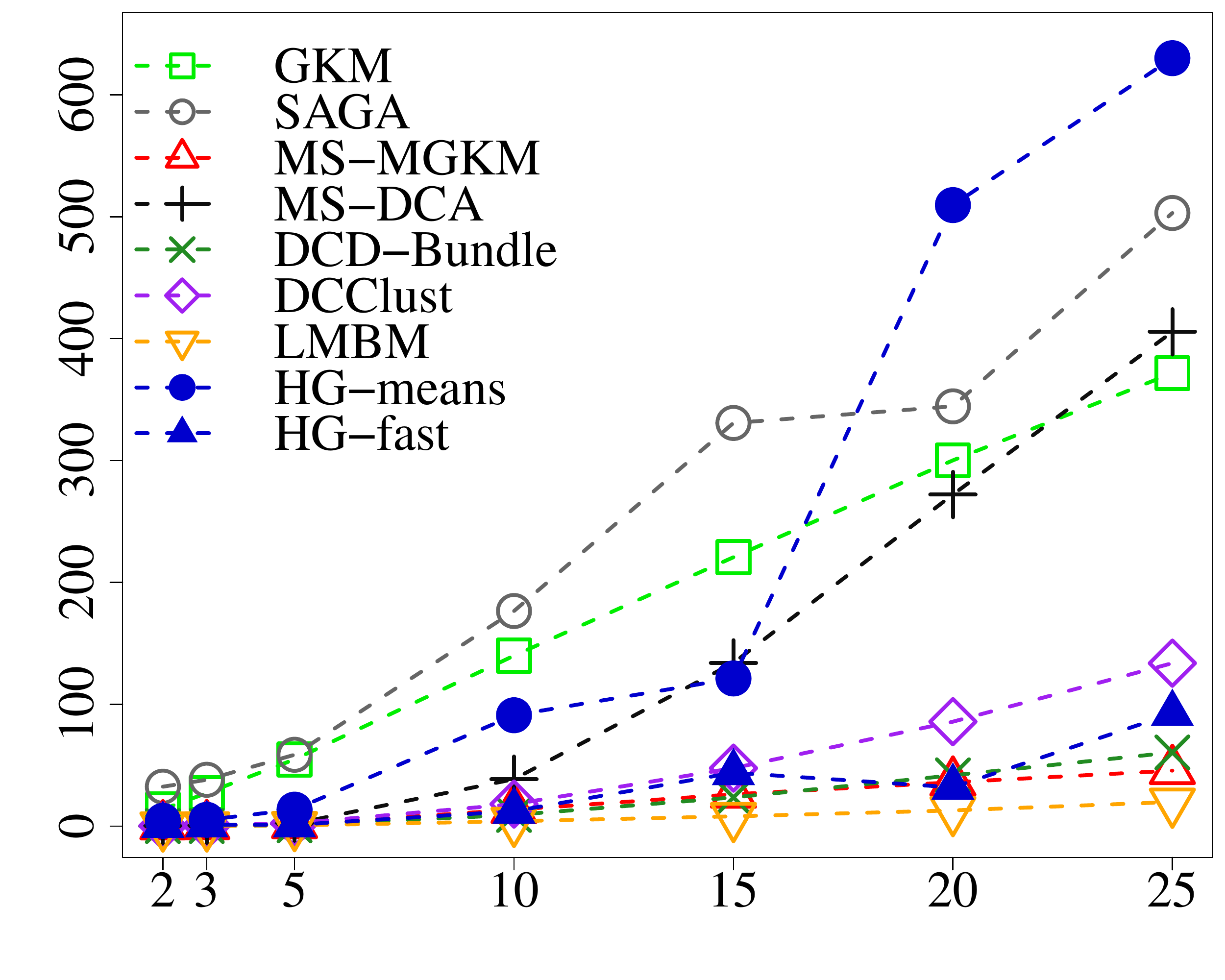}
\end{minipage}
\hspace*{0cm}

\begin{minipage}{0.322\textwidth}
\scalebox{0.9}{\hspace*{0.55cm}Shuttle ($n = 58000$, $d = 9$)}
\vspace*{0.08cm}
\includegraphics[width=\textwidth]{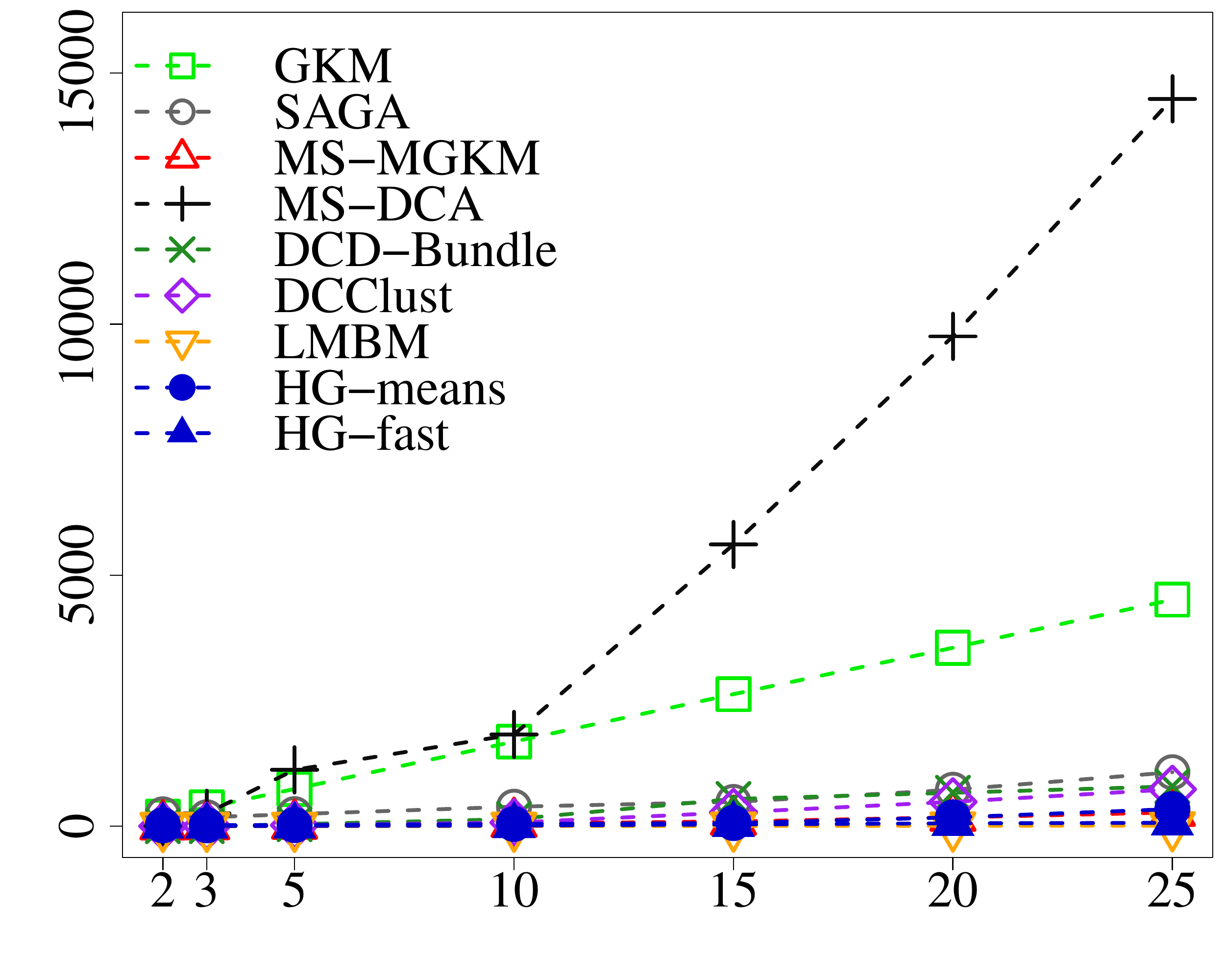}
\end{minipage}
\vspace*{0.0cm}
\begin{minipage}{0.322\textwidth}
\scalebox{0.9}{\hspace*{0.55cm}Skin ($n = 245057$, $d = 3$)}
\vspace*{0.08cm}
\includegraphics[width=\textwidth]{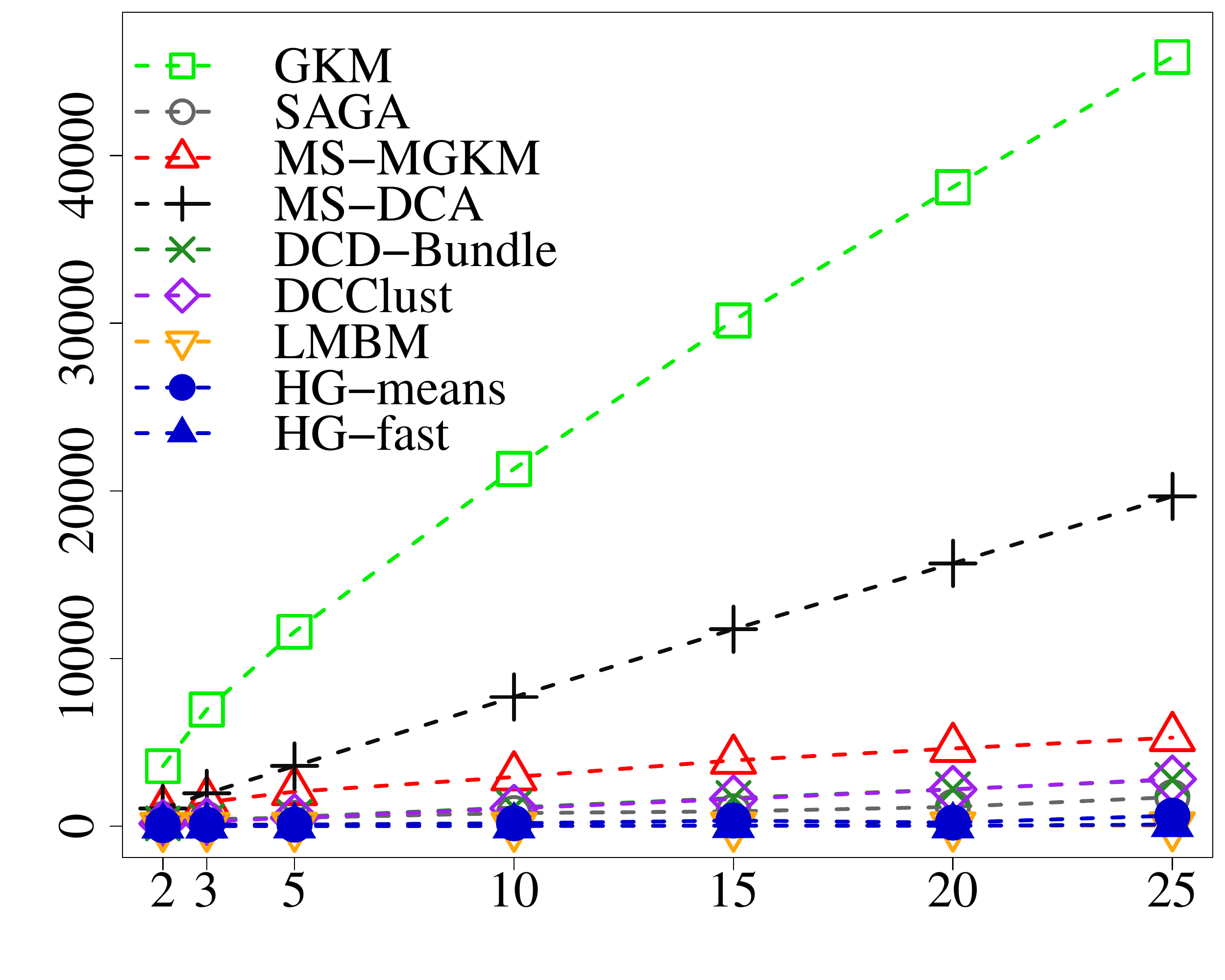}
\end{minipage}
\hspace*{0cm}
\begin{minipage}{0.322\textwidth}
\scalebox{0.9}{\hspace*{0.55cm}KEGG ($n = 53413$, $d = 20$)}
\vspace*{0.08cm}
\includegraphics[width=\textwidth]{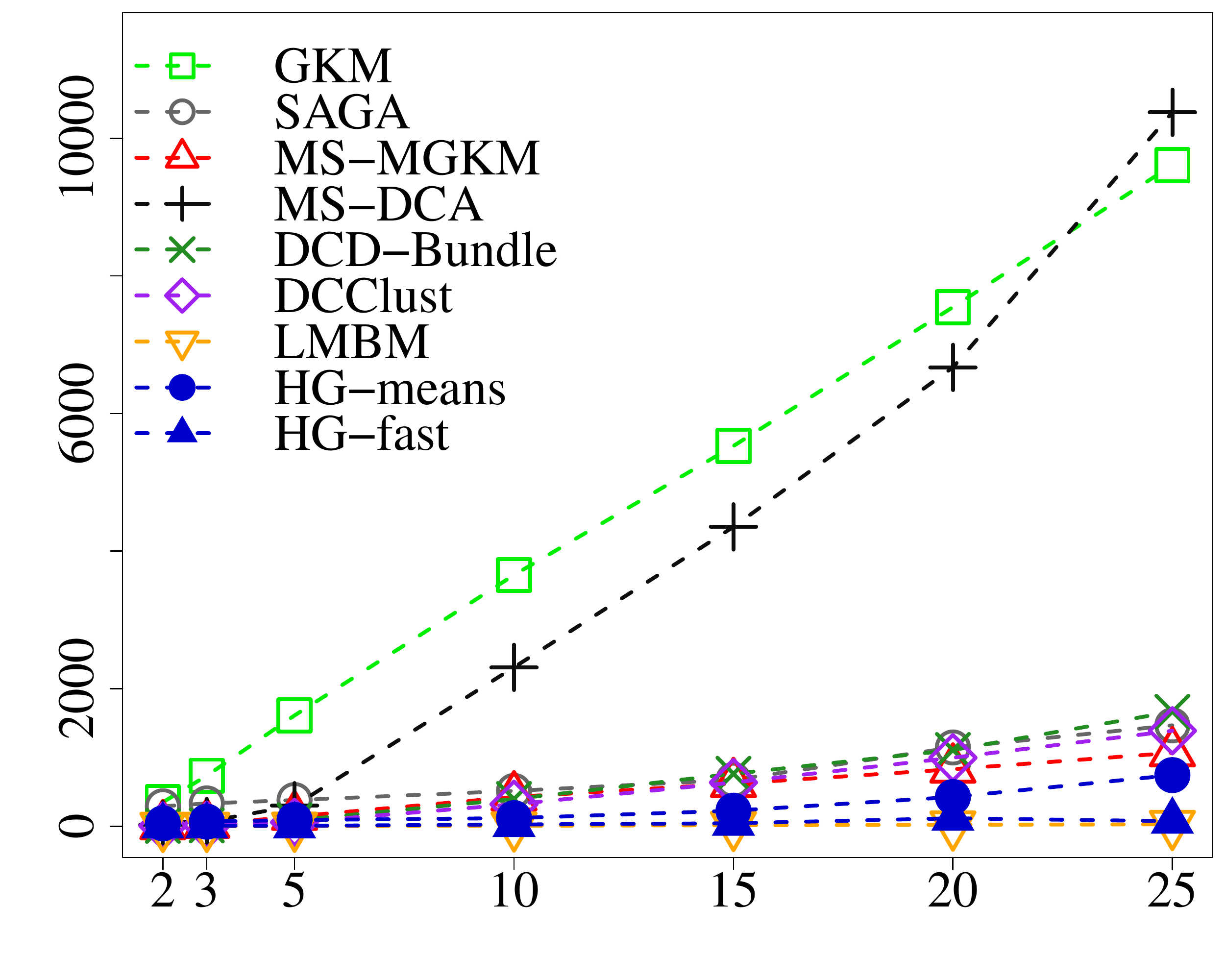}
\end{minipage}
\hspace*{0cm}

\begin{minipage}{0.322\textwidth}
\scalebox{0.9}{\hspace*{0.55cm}3D road ($n = 434874$, $d = 3$)}
\vspace*{0.08cm}
\includegraphics[width=\textwidth]{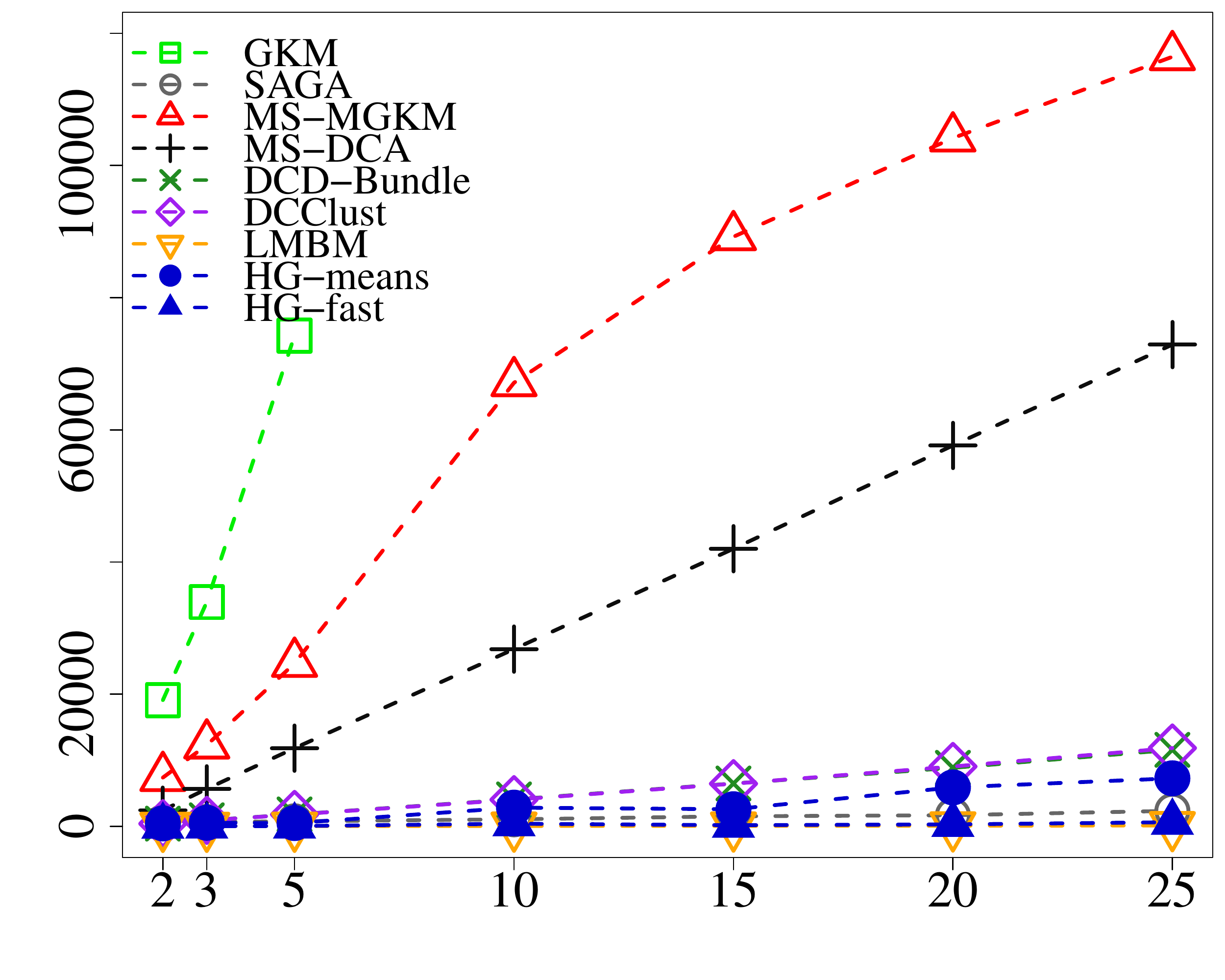}
\end{minipage}
\vspace*{0.0cm}
\begin{minipage}{0.322\textwidth}
\scalebox{0.9}{\hspace*{0.55cm}Gas sensor ($n = 13910$, $d = 128$)}
\vspace*{0.08cm}
\includegraphics[width=\textwidth]{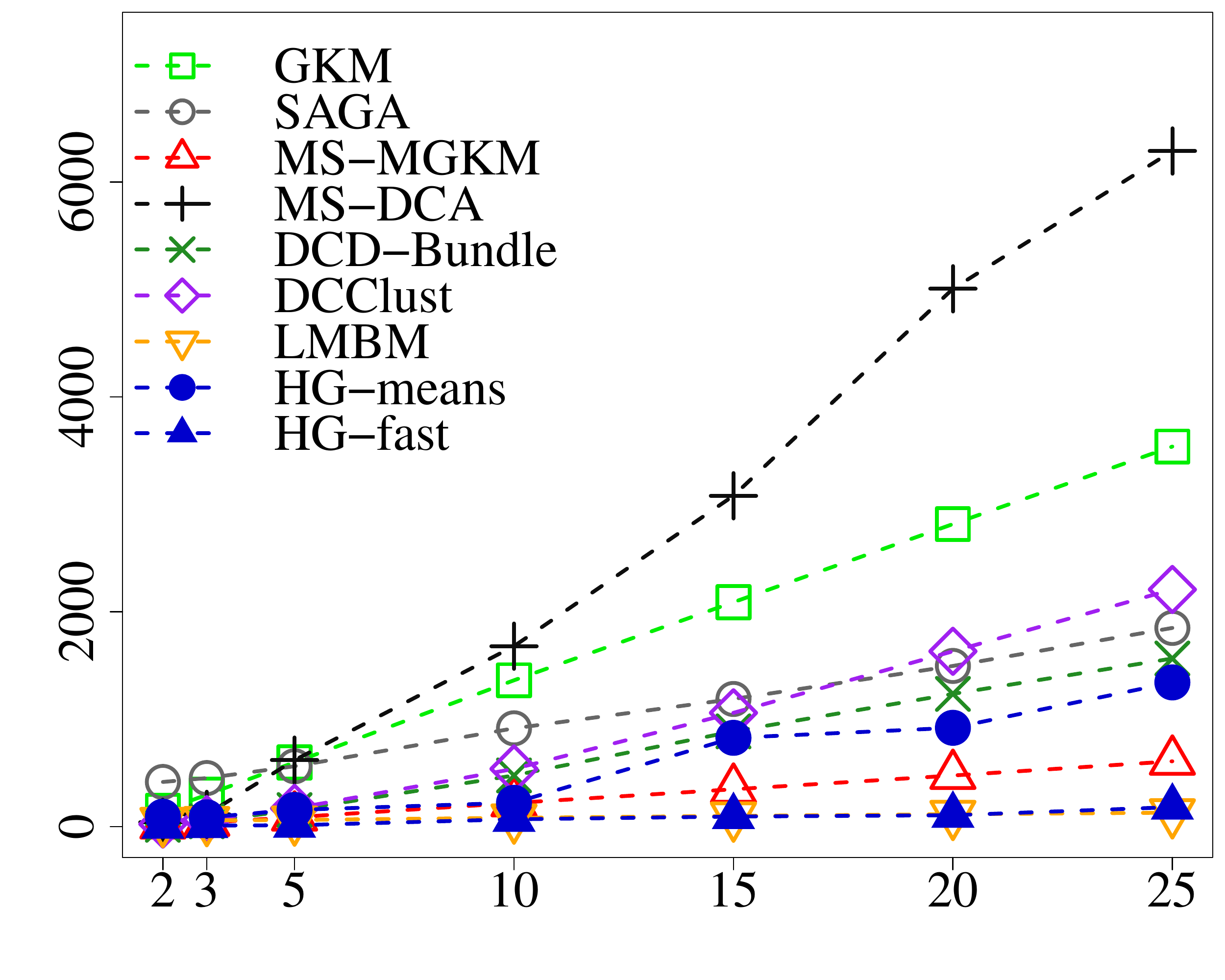}
\end{minipage}
\hspace*{0cm}
\begin{minipage}{0.322\textwidth}
\scalebox{0.9}{\hspace*{0.55cm}Online News ($n = 39644$, $d = 58$)}
\vspace*{0.08cm}
\includegraphics[width=\textwidth]{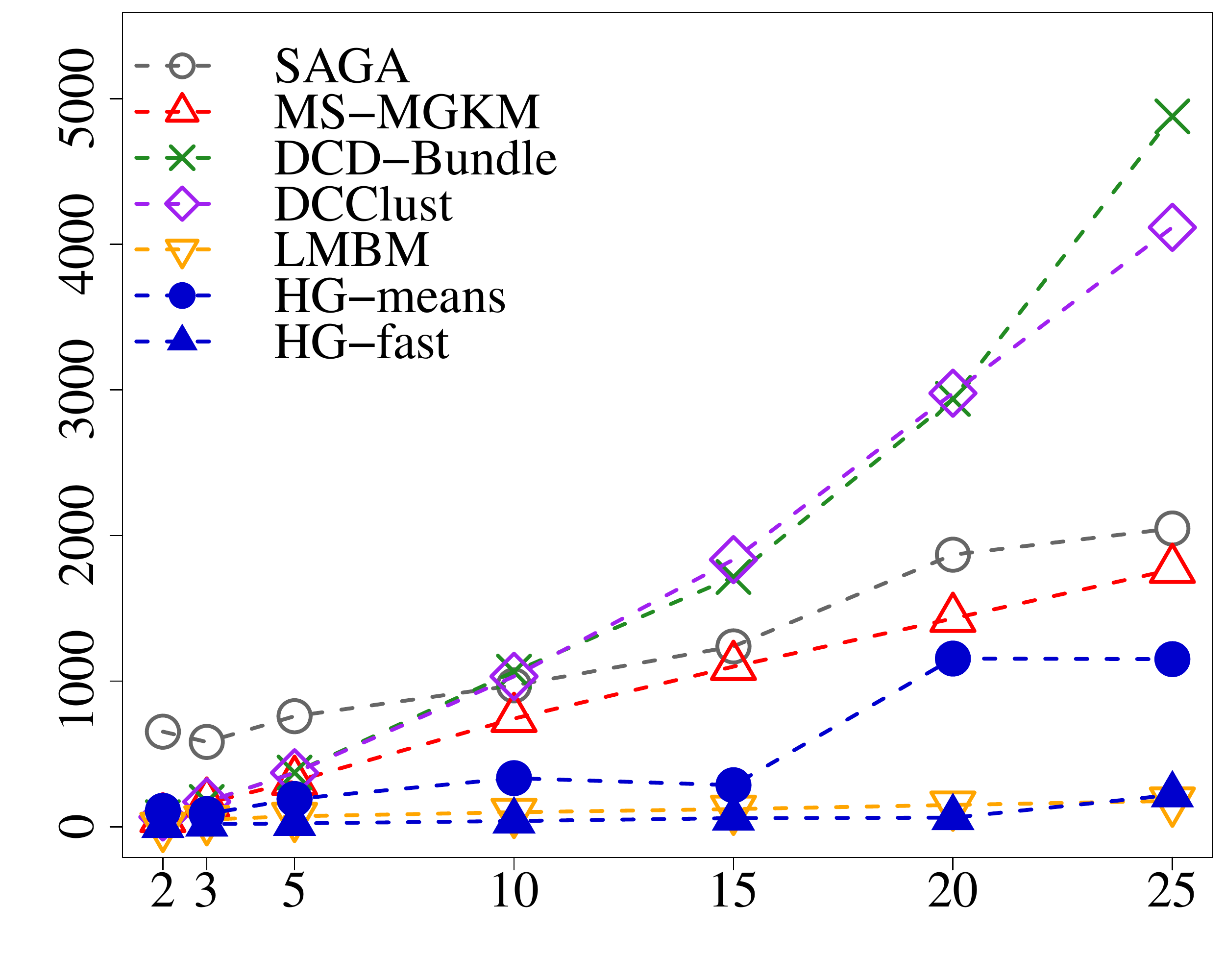}
\end{minipage}
\hspace*{0cm}

\begin{minipage}{0.322\textwidth}
\scalebox{0.9}{\hspace*{0.55cm}Sensorless ($n = 58509$, $d = 48$)}
\vspace*{0.08cm}
\includegraphics[width=\textwidth]{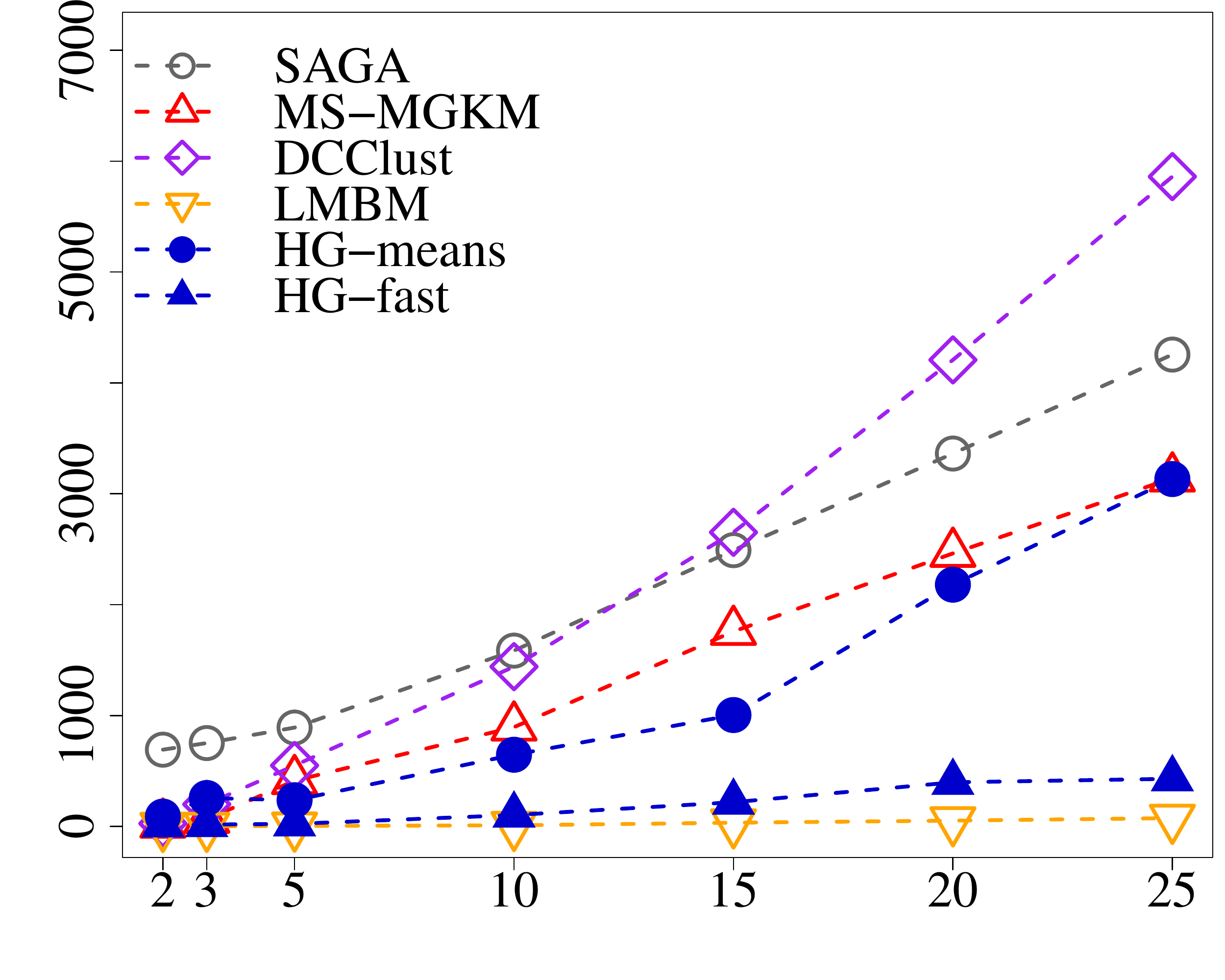}
\end{minipage}
\vspace*{0.0cm}
\begin{minipage}{0.322\textwidth}
\scalebox{0.9}{\hspace*{0.55cm}Isolet ($n = 7797$, $d = 617$)}
\vspace*{0.08cm}
\includegraphics[width=\textwidth]{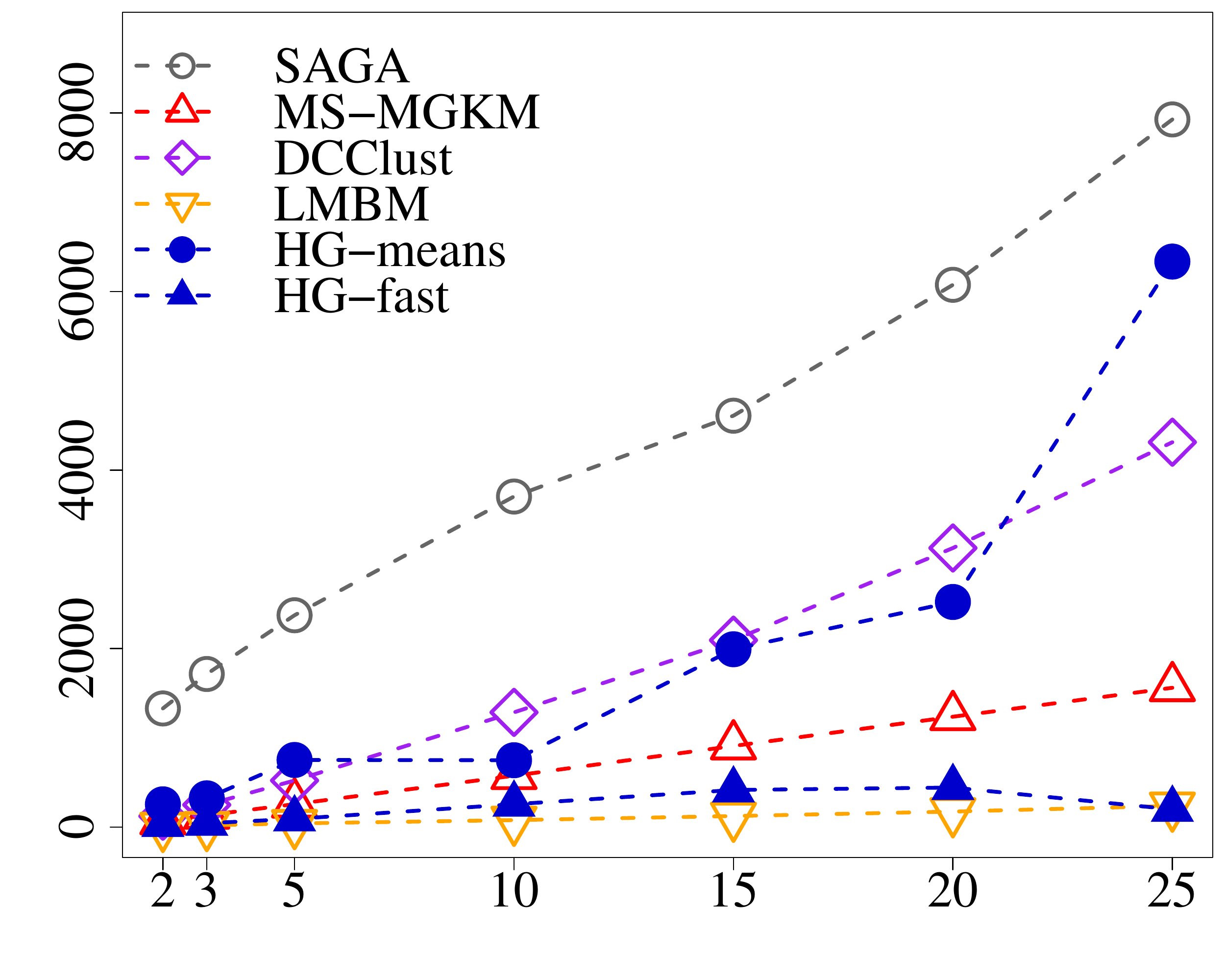}
\end{minipage}
\hspace*{0cm}

\begin{minipage}{0.322\textwidth}
\scalebox{0.9}{\hspace*{0.55cm}MiniBooNE ($n = 130064$, $d = 50$)}
\vspace*{0.08cm}
\includegraphics[width=\textwidth]{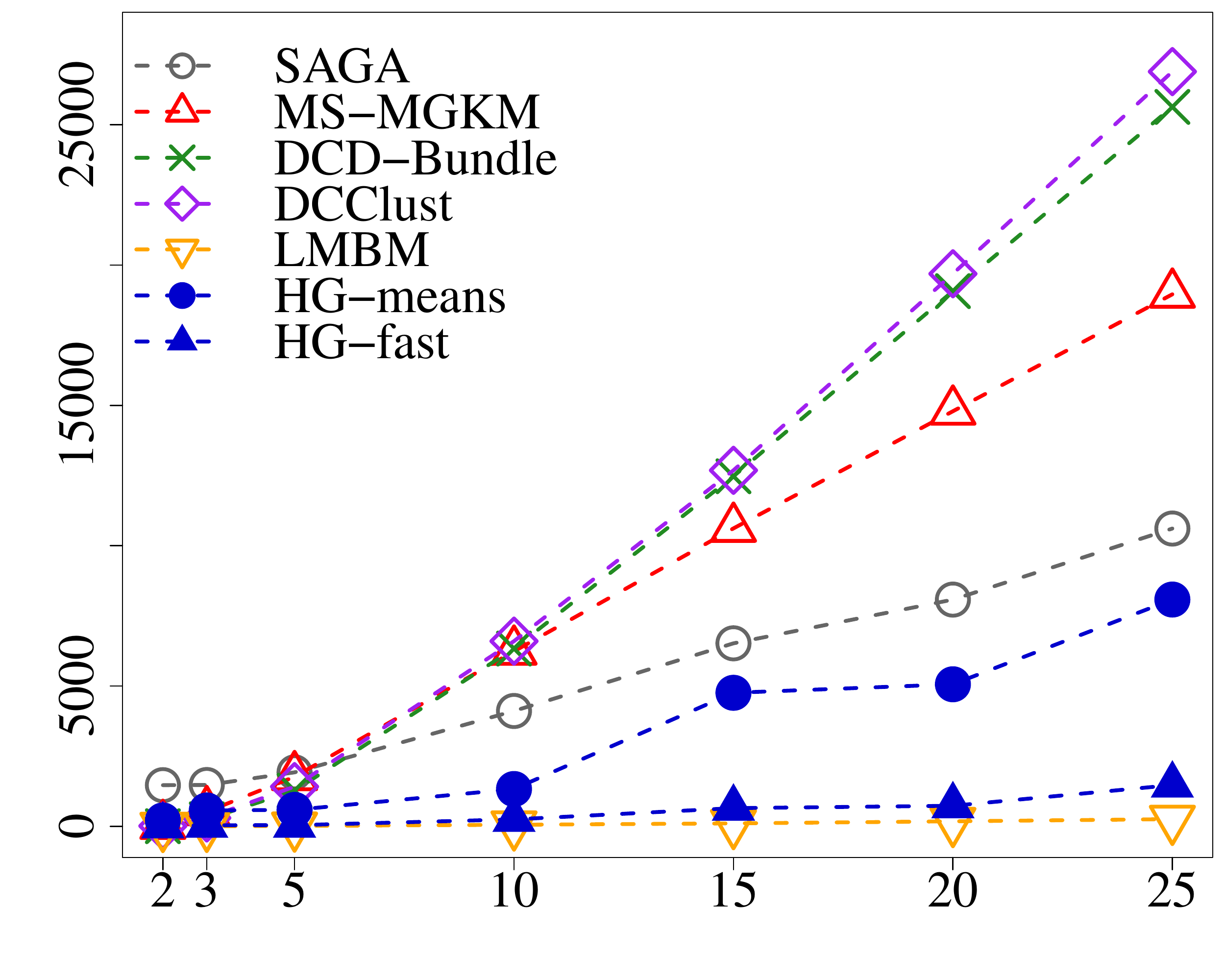}
\end{minipage}
\hspace*{0cm}
\begin{minipage}{0.322\textwidth}
\scalebox{0.9}{\hspace*{0.55cm}Gisette ($n = 13500$, $d = 5000$)}
\vspace*{0.08cm}
\includegraphics[width=\textwidth]{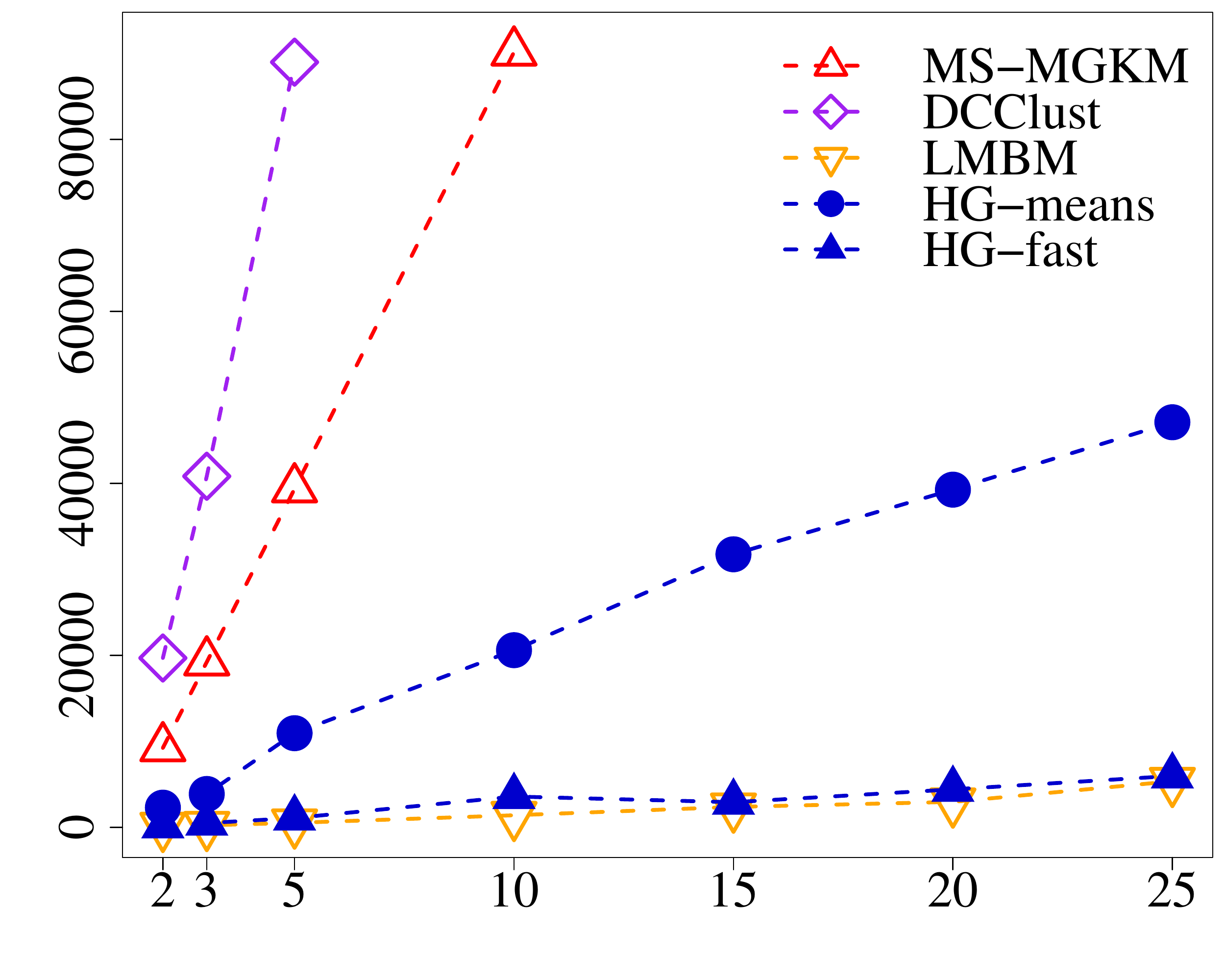}
\end{minipage}
\hspace*{0cm}
\vspace*{-0.5cm}

\caption{CPU time of state-of-the-art algorithms on class C datasets, as a function of the number of clusters}
\label{fig:timeC}
\end{figure}

We observe a significant reduction in CPU time when comparing the results of the \emph{fast} \textsc{HG-means} in Table \ref{table:detailed:fast} with those of the standard version in Table \ref{table:detailed}.
Considering the speed ratio between methods for each dataset, the fast configuration is in average seven times faster than the standard \textsc{HG-means} and over 10 times faster than SAGA, MS-MGKM, DCClust, DCD-Bundle and MS-DCA. Figure \ref{fig:timeC} also displays the CPU time of \textsc{HG-means}, its \emph{fast} configuration, and the other algorithms listed in Section~\ref{sec:results} as a function of $m$.
Surprisingly, the solution quality did not deteriorate much by reducing the termination criterion for these large datasets: with a percentage gap of $-0.25\%$, the solutions found by the fast \textsc{HG-means} are close to those of the standard version (gap of $-0.26\%$) and still much better than all solutions found in previous studies. Therefore, researchers interested in using \textsc{HG-means} can easily adapt the termination criterion of the method, so as to obtain significant gains of solution quality within their own computational budget.


\subsection{Solution Quality and \myblue{Clustering} Performance}
\label{sec:gaussian}

The previous section has established that \textsc{HG-means} finds better MSSC local minima than other state-of-the-art algorithms and exemplified the known fact that \textsc{K-means} or \textsc{K-means++} solutions can be arbitrarily far from the global minimum. In most situations, using the most accurate method for a problem should be the best option. However, \textsc{HG-means} is slower than a few runs of \textsc{K-means} or \textsc{K-means++}. To determine whether it is worth investing this additional effort, we must determine whether a better solution quality for the MSSC problem effectively translates into better \myblue{clustering} performance. There have been similar investigations in other machine learning subfields, e.g., to choose the amount of effort dedicated to training deep neural networks (see, e.g. \citep{Hoffer2017,Keskar2017}).

To explore this, we conduct an experiment in which we compare the ability of \textsc{HG-means}, \textsc{K-means} and \textsc{K-means++} to classify 50,000 samples issued from a mixture of spherical Gaussian distributions: $X \sim \nicefrac{1}{m} \sum_{i=1}^m \mathcal{N}(\boldsymbol\mu_i,\boldsymbol\Sigma_i)$ with $\boldsymbol\Sigma_i = \sigma^2_i \mathbf{I}$. For each $i \in \{1,\dots,m\}$, $\mu_i$ and $\sigma^2_i$ are uniformly selected in
 $[0,5]$ and $[1,10]$, respectively. This is a fundamental setting, without any hidden structure, in which we expect the MSSC model and the associated \textsc{K-means} variants to be a good choice since these methods promote spherical and balanced clusters. To increase the challenge, we consider a medium to large number of Gaussians, with $m \in \{20,50,100,200\}$, in feature spaces of medium to high dimensions $d \in \{20,50,100,200,500\}$. For each combination of $m$ and $d$, we repeat the generation process until we obtain a mixture that is not $1$-separated and in which at least 99\% of the pairs of Gaussians are $\nicefrac{1}{2}$-separated \citep{Dasgupta1999}. Such a mixture corresponds to Gaussians that significantly overlap. These datasets can be accessed at \url{https://w1.cirrelt.ca/~vidalt/en/research-data.html}.

Tables~\ref{Gauss-all1} and~\ref{Gauss-all2} compare the results of \textsc{HG-means} with those of \textsc{K-means} and \textsc{K-means++} over a single run or 500 repeated runs, in terms of MSSC solution quality (as represented by the percentage gap) and \myblue{cluster validity} in relation to the ground truth.
We use three \myblue{external measures of cluster validity}: the adjusted Rand index (CRand -- \cite{Hubert1985}), the normalized mutual information (NMI -- \cite{Kvalseth1987}), and the centroid index (CI -- \cite{Franti2014}). CRand and NMI take continuous values in $[-1,1]$ and $[0,1]$, respectively. They converge toward $1$ as the clustering solution becomes closer to the ground truth. CI takes integer values and measures the number of fundamental cluster differences between solutions, with a value of $0$ indicating that the solution has the same cluster-level structure as the ground truth.

As in the previous experiments, a single run of \textsc{K-means} or \textsc{K-means++} visibly leads to shallow local minima that can be improved with multiple runs from different starting points. However, even 500 repetitions of these algorithms are insufficient to reach the solution quality of \textsc{HG-means}, despite the similar CPU time.
\textsc{K-means} performs better than \textsc{K-means++} for these datasets, most likely because it is more robust to outliers when selecting initial center locations. A pairwise Wilcoxon test highlights significant differences between \textsc{HG-means} and all other methods (p-values $\leq 0.0002$). With a Pearson coefficient $r \geq 0.8$, the dimension $d$ of the feature space is correlated to the inaccuracy (percentage gap) of the repeated \textsc{K-means} and \textsc{K-means++} algorithms, which appear to be more easily trapped in low-quality local minima for feature spaces of larger dimension.

Comparing the external \myblue{clustering} performance of these methods (via CRand, NMI, and CI) leads to additional insights.
For all three metrics, pairwise Wilcoxon tests highlight significant performance differences between \textsc{HG-means} and repeated \textsc{K-means} variants (with p-values $\leq  3.1 \times 10^{-5}$).
We also observe a correlation between the solution quality (percentage gap) and the three external measures.
Although the differences in the MSSC objective function values appear small at first glance (e.g., average gap of 4.87\% for repeated \textsc{K-means}), these inaccuracies have a large effect on \myblue{cluster validity}, especially for datasets with feature spaces of higher dimension.
When $d=500$, \textsc{HG-means} is able to exploit the increased amount of available independent information to find a close approximation to the ground truth (average CRand of 0.99, NMI of 1.00, and CI of 1.75) whereas repeated \textsc{K-means} and \textsc{K-means++} reach shallow local optima and obtain inaccurate results (average CRand and NMI below 0.65 and 0.94, respectively, and average CI above 23.75).
Classical distance metrics are known to become more uniform as the feature-space dimension grows, and the number of local minima of the MSSC quickly increases, so feature-reduction or subspace-clustering techniques are often recommended for high-dimensional datasets. In these experiments, however, the inaccuracy of repeated \textsc{K-means} (or \textsc{K-means++}) appears to be a direct consequence of its inability to find good-quality local minima, rather than a consequence of the MSSC model itself, since near-optimal solutions of the MSSC translate into accurate results.

Overall, we conclude that even for simple Gaussian-based distribution mixtures, finding good local minima of the MSSC problem is essential for an accurate information retrieval. This is a major difference with studies on, for example, deep neural networks, where it is conjectured that most local minima have similar objective values, and where more intensive training (e.g., stochastic gradient descent with large batches) have adverse effects on generalization \cite{Lecun2015}. 
For clustering problems, it is essential to keep progressing toward faster and more accurate MSSC solvers, and to recognize when to use these high-performance methods for large and high-dimensional datasets.

\begin{landscape}
\begin{table}[htbp]
\vspace*{-1.0cm}
\centering
\renewcommand{\arraystretch}{1.02}
{
\setlength\tabcolsep{9pt}
\scalebox{0.75}{
\begin{tabular}{|cc|c|ccccc|ccccc|cccc}
\hline
&&BKS&\multicolumn{5}{c|}{\textbf{Gap (\%)}}&\multicolumn{5}{c|}{\textbf{Time (s)}}\\
\multicolumn{2}{|c|}{\strut}&Objective&\multicolumn{2}{c}{\textsc{K-means}}&\multicolumn{2}{c}{\textsc{K-means++}}&\textsc{HG-means}&\multicolumn{2}{c}{\textsc{K-means}}&\multicolumn{2}{c}{\textsc{K-means++}}&\textsc{HG-means}\\
m&d&Value&1 Run&500 Runs&1 Run&500 Runs&&1 Run&500 Runs&1 Run&500 Runs&\\
\hline
20&20&5432601.91&0.73&\textbf{0.0}&1.15&\textbf{0.0}&\textbf{0.0}&2.40&668.93&3.00&764.76&1085.40\\
20&50&12815114.52&6.19&\textbf{0.0}&3.75&1.15&\textbf{0.0}&2.86&860.95&3.17&1171.09&1308.96\\
20&100&24266784.28&14.84&\textbf{0.0}&5.01&4.83&\textbf{0.0}&5.38&1243.53&4.75&1958.25&553.25\\
20&200&59340268.17&17.70&2.57&11.79&7.00&\textbf{0.0}&14.90&2677.57&12.43&3938.29&1505.16\\
20&500&125359202.26&16.53&8.06&25.35&8.00&\textbf{0.0}&30.13&6118.59&25.17&8389.50&2563.73\\
50&20&5305274.24&0.47&\textbf{0.0}&0.43&\textbf{0.0}&\textbf{0.0}&5.03&2599.11&4.84&2755.17&3189.56\\
50&50&13864882.54&2.10&\textbf{0.0}&3.22&0.72&\textbf{0.0}&7.28&2695.69&8.23&3258.11&4307.12\\
50&100&25645070.92&8.86&3.70&12.04&5.76&\textbf{0.0}&10.78&4226.64&14.33&5871.70&2934.41\\
50&200&52561077.57&14.62&7.76&19.90&9.92&\textbf{0.0}&22.98&7837.70&37.60&11063.60&9629.09\\
50&500&143469250.17&16.92&9.79&20.0&11.11&\textbf{0.0}&38.89&14778.04&58.13&19077.48&18360.24\\
100&20&5027688.54&0.34&0.12&0.54&0.04&\textbf{0.0}&19.79&7281.83&18.89&8435.48&13529.09\\
100&50&12897680.57&3.07&1.17&4.81&2.25&\textbf{0.0}&12.07&6612.89&15.27&7962.07&10344.57\\
100&100&27284752.32&6.30&4.67&10.58&6.89&\textbf{0.0}&24.43&11864.87&30.54&14991.49&6728.71\\
100&200&51552765.51&14.03&7.97&15.78&11.13&\textbf{0.0}&34.63&14537.27&52.73&20128.89&20499.22\\
100&500&130903680.95&18.90&15.61&22.71&15.69&\textbf{0.0}&61.45&25313.95&86.04&34062.29&38217.57\\
200&20&4774890.45&0.72&0.45&1.24&0.53&\textbf{0.0}&42.85&18861.45&38.91&19896.36&38126.21\\
200&50&13490838.00&1.97&1.16&2.88&1.86&\textbf{0.0}&34.49&18792.14&39.88&21036.63&28513.22\\
200&100&27337380.17&8.08&5.29&9.68&7.56&\textbf{0.0}&70.30&30880.39&82.03&36219.66&39980.98\\
200&200&52946223.09&15.77&11.70&19.67&14.45&\textbf{0.0}&74.33&37459.76&139.62&46365.94&67745.79\\
200&500&135201463.76&20.97&17.32&23.83&19.28&\textbf{0.0}&142.85&63202.41&210.16&92765.62&93444.51\\
\hline
\end{tabular}
}}
\caption{Mixture of spherical Gaussian distributions -- Solution quality and CPU time}
\label{Gauss-all1}
\vspace{0.75cm}
\scalebox{0.75}{
\begin{tabular}{|cc|ccccc|ccccc|ccccc|}
\hline
&&\multicolumn{5}{c|}{\textbf{CRand}}&\multicolumn{5}{c|}{\textbf{NMI}}&\multicolumn{5}{c|}{\textbf{CI}}\\
\multicolumn{2}{|c|}{\strut}&\multicolumn{2}{c}{\textsc{K-means}}&\multicolumn{2}{c}{\textsc{K-means++}}&{\textsc{HG-means}}&\multicolumn{2}{c}{\textsc{K-means}}&\multicolumn{2}{c}{\textsc{K-means++}}&{\textsc{HG-means}}&\multicolumn{2}{c}{\textsc{K-means}}&\multicolumn{2}{c}{\textsc{K-means++}}&{\textsc{HG-means}}\\
m&d&1 Run&500 Runs&1 Run&500 Runs&&1 Run&500 Runs&1 Run&500 Runs&&1 Run&500 Runs&1 Run&500 Runs&\\
\hline
20&20&0.69&\textbf{0.72}&0.67&\textbf{0.72}&\textbf{0.72}&0.73&\textbf{0.75}&0.73&\textbf{0.75}&\textbf{0.75}&1&\textbf{0}&1&\textbf{0}&\textbf{0}\\
20&50&0.76&\textbf{0.98}&0.86&0.92&\textbf{0.98}&0.91&\textbf{0.98}&0.94&0.96&\textbf{0.98}&3&\textbf{0}&2&1&\textbf{0}\\
20&100&0.63&\textbf{1.00}&0.89&0.89&\textbf{1.00}&0.89&\textbf{1.00}&0.97&0.97&\textbf{1.00}&5&\textbf{0}&2&2&\textbf{0}\\
20&200&0.47&0.94&0.61&0.83&\textbf{1.00}&0.84&0.98&0.89&0.95&\textbf{1.00}&7&1&5&3&\textbf{0}\\
20&500&0.55&0.81&0.32&0.81&\textbf{1.00}&0.88&0.95&0.79&0.95&\textbf{1.00}&6&2&9&3&\textbf{0}\\
50&20&0.58&\textbf{0.59}&0.57&\textbf{0.59}&\textbf{0.59}&0.67&\textbf{0.68}&0.67&\textbf{0.68}&\textbf{0.68}&1&\textbf{0}&2&\textbf{0}&\textbf{0}\\
50&50&0.87&\textbf{0.94}&0.82&0.92&\textbf{0.94}&0.93&\textbf{0.95}&0.92&0.94&\textbf{0.95}&3&\textbf{0}&5&1&\textbf{0}\\
50&100&0.76&0.90&0.59&0.85&\textbf{1.00}&0.95&0.98&0.92&0.96&\textbf{1.00}&9&4&12&6&\textbf{0}\\
50&200&0.52&0.80&0.34&0.72&\textbf{1.00}&0.90&0.96&0.85&0.94&\textbf{1.00}&14&8&19&10&\textbf{0}\\
50&500&0.41&0.69&0.24&0.39&\textbf{1.00}&0.88&0.94&0.83&0.91&\textbf{1.00}&16&9&16&10&\textbf{0}\\
100&20&0.48&0.48&0.47&\textbf{0.49}&\textbf{0.49}&0.62&\textbf{0.63}&0.62&\textbf{0.63}&\textbf{0.63}&4&2&5&1&\textbf{0}\\
100&50&0.80&0.86&0.78&0.84&\textbf{0.91}&0.91&0.93&0.90&0.92&\textbf{0.94}&9&4&13&6&\textbf{0}\\
100&100&0.80&0.86&0.68&0.74&\textbf{0.99}&0.96&0.97&0.93&0.94&\textbf{1.00}&15&11&23&16&\textbf{1}\\
100&200&0.63&0.79&0.53&0.74&\textbf{0.99}&0.93&0.96&0.92&0.95&\textbf{1.00}&27&16&30&20&\textbf{1}\\
100&500&0.40&0.60&0.23&0.35&\textbf{0.98}&0.89&0.93&0.84&0.90&\textbf{1.00}&33&27&37&29&\textbf{2}\\
200&20&0.39&0.40&0.38&0.39&\textbf{0.41}&0.59&0.59&0.58&0.59&\textbf{0.60}&22&14&25&20&\textbf{6}\\
200&50&0.81&0.82&0.78&0.80&\textbf{0.87}&0.91&0.90&0.90&0.89&\textbf{0.92}&12&10&18&13&\textbf{0}\\
200&100&0.71&0.81&0.66&0.73&\textbf{0.96}&0.94&0.95&0.94&0.94&\textbf{0.99}&38&27&49&38&\textbf{5}\\
200&200&0.51&0.64&0.31&0.56&\textbf{0.99}&0.92&0.94&0.87&0.93&\textbf{1.00}&61&45&71&53&\textbf{3}\\
200&500&0.41&0.50&0.26&0.33&\textbf{0.98}&0.90&0.92&0.85&0.89&\textbf{1.00}&65&57&74&60&\textbf{5}\\
\hline
\end{tabular}
}
\caption{Mixture of spherical Gaussian distributions -- \myblue{External cluster validity}}
\label{Gauss-all2}
\end{table}
\end{landscape}

\section{Conclusions and Perspectives}
\label{sec:conclusions}

In this article, we have studied the MSSC problem, a classical clustering model of which the popular \textsc{K-means} algorithm constitutes a local minimizer.
We have proposed a hybrid genetic algorithm, \textsc{HG-means}, that combines the improvement capabilities of \textsc{K-means} as a local search with the diversification capabilities of problem-tailored genetic operators. The algorithm uses an exact minimum-cost matching crossover operator and an adaptive mutation procedure to generate strategic initial center positions for \textsc{K-means} and to promote a thorough exploration of the search space. Moreover, it uses population diversity management strategies to prevent premature convergence to shallow local minima. 

We conducted extensive computational experiments to evaluate the performance of the method in terms of MSSC solution quality, computational effort and scalability, and \myblue{external cluster validity}. Our results indicate that \textsc{HG-means} attains better local minima than all recent state-of-the-art algorithms. Large solution improvements are usually observed for large datasets with a medium-to-high number of clusters $m$, since these characteristics lead to MSSC problems that have not been effectively solved by previous approaches. The CPU time of \textsc{HG-means} is directly proportional to that of the \textsc{K-means} local-improvement procedure and to the number of iterations allowed (the termination criterion). It appears to grow linearly with the number of samples and feature-space dimensions, and the termination criterion can be adjusted to achieve solutions in a shorter time without a large impact on solution accuracy.

Through additional tests conducted on datasets generated via Gaussian mixtures, we observed a strong correlation between MSSC solution quality and \myblue{cluster validity measures}. A repeated \textsc{K-means} algorithm, for example, obtains solution inaccuracies (percentage gap to the best known local minima) that are small at first glance but highly detrimental for the outcome of the \myblue{clustering task}. In particular, a gap as small as 5\% in the objective space can make the difference between accurate \myblue{clustering} and failure. This effect was observed in all Gaussian datasets studied, and it became more prominent in feature spaces of higher dimension. In those situations, the inability of \textsc{K-means} or \textsc{K-means++} to provide satisfactory results seems to be tied to its inability to find good-quality local minima of the MSSC model, rather than to an inadequacy of the model itself.

Overall, beyond the immediate gains in terms of \myblue{clustering} performance, research into efficient optimization algorithms for MSSC remains linked to important methodological stakes. Indeed, a number of studies aim to find adequate models (e.g., MSSC) for different tasks and datasets. With that goal in mind, it is essential to differentiate the limitations of the model themselves (inadequacy for a given task or data type), and those of algorithms used to solve such models (shallow local optima). While an analysis using external measures (e.g., CRand, NMI or CI) allows a general evaluation of error (due to both sources of inaccuracy), only a precise investigation of a method's performance in the objective space can help evaluating the magnitude of each imprecision, and only accurate or exact optimization methods can give meaningful conclusions regarding model suitability. In future research, we plan to continue progressing in this direction, generalizing the proposed solution method to other clustering models, possibly considering the use of kernel transformations, integrating semi-supervised information in the form of must-link or cannot-link constraints, and pursuing the development of high-performance optimization algorithms for other classes of applications.

\section{Acknowledgments}
The authors thank the four anonymous referees for their detailed comments, which significantly contributed to improving this paper. This research is partially supported by CAPES, CNPq [grant number 308498/2015-1] and FAPERJ [grant number E-26/203.310/2016] in Brazil.
This support is gratefully acknowledged.


\begin{thebibliography}{43}
\expandafter\ifx\csname natexlab\endcsname\relax\def\natexlab#1{#1}\fi
\expandafter\ifx\csname url\endcsname\relax
  \def\url#1{{\tt #1}}\fi
\expandafter\ifx\csname urlprefix\endcsname\relax\def\urlprefix{URL }\fi
\expandafter\ifx\csname urlstyle\endcsname\relax
  \expandafter\ifx\csname doi\endcsname\relax
  \def\doi#1{doi:\discretionary{}{}{}#1}\fi \else
  \expandafter\ifx\csname doi\endcsname\relax
  \def\doi{doi:\discretionary{}{}{}\begingroup \urlstyle{rm}\Url}\fi \fi

\bibitem[{Al-Sultan(1995)}]{Al-Sultan1995}
Al-Sultan, K. 1995.
\newblock {A tabu search approach to the clustering problem}.
\newblock {\it Pattern Recognition\/} {\bf 28}(9) 1443--1451.

\bibitem[{Aloise et~al.(2009)Aloise, Deshpande, Hansen, and
  Popat}]{Aloise2009a}
Aloise, D., A.~Deshpande, P.~Hansen, P.~Popat. 2009.
\newblock {NP-hardness of Euclidean sum-of-squares clustering}.
\newblock {\it Machine Learning\/} {\bf 75}(2) 245--248.

\bibitem[{Aloise et~al.(2012)Aloise, Hansen, and Liberti}]{Aloise2012exact}
Aloise, D., P.~Hansen, L.~Liberti. 2012.
\newblock {An improved column generation algorithm for minimum sum-of-squares
  clustering}.
\newblock {\it Mathematical Programming\/} {\bf 131}(1) 195--220.

\bibitem[{An et~al.(2014)An, Minh, and Tao}]{HoaiAn2014}
An, L.T.H., L.H. Minh, P.D. Tao. 2014.
\newblock {New and efficient DCA based algorithms for minimum sum-of-squares
  clustering}.
\newblock {\it Pattern Recognition\/} {\bf 47}(1) 388--401.

\bibitem[{Arthur and Vassilvitskii(2007)}]{Arthur2007}
Arthur, D., S.~Vassilvitskii. 2007.
\newblock {K-Means++: The advantages of careful seeding}.
\newblock {\it SODA'07. Proceedings of the Eighteenth Annual ACM-SIAM Symposium
  on Discrete Algorithms\/}. SIAM, New Orleans, Louisiana, USA, 1027--1035.

\bibitem[{Bagirov(2008)}]{Bagirov2008}
Bagirov, A.M. 2008.
\newblock {Modified global k-means algorithm for minimum sum-of-squares
  clustering problems}.
\newblock {\it Pattern Recognition\/} {\bf 41}(10) 3192--3199.

\bibitem[{Bagirov et~al.(2016)Bagirov, Taheri, and Ugon}]{Bagirov2016}
Bagirov, A.M., S.~Taheri, J.~Ugon. 2016.
\newblock {Nonsmooth DC programming approach to the minimum sum-of-squares
  clustering problems}.
\newblock {\it Pattern Recognition\/} {\bf 53} 12--24.

\bibitem[{Bagirov et~al.(2011)Bagirov, Ugon, and Webb}]{Bagirov2011}
Bagirov, A.M., J.~Ugon, D.~Webb. 2011.
\newblock {Fast modified global k-means algorithm for incremental cluster
  construction}.
\newblock {\it Pattern Recognition\/} {\bf 44}(4) 866--876.

\bibitem[{Blum et~al.(2011)Blum, Puchinger, Raidl, and Roli}]{Blum2011}
Blum, C., J.~Puchinger, G.~Raidl, A.~Roli. 2011.
\newblock {Hybrid metaheuristics in combinatorial optimization: A survey}.
\newblock {\it Applied Soft Computing\/} {\bf 11}(6) 4135--4151.

\bibitem[{Dasgupta(1999)}]{Dasgupta1999}
Dasgupta, S. 1999.
\newblock {Learning mixtures of Gaussians}.
\newblock {\it 40th Annual Symposium on Foundations of Computer Science\/} {\bf
  1} 634--644.

\bibitem[{Fr{\"{a}}nti et~al.(1997)Fr{\"{a}}nti, Kivij{\"{a}}rvi, Kaukoranta,
  and Nevalainen}]{Franti1997}
Fr{\"{a}}nti, P., J.~Kivij{\"{a}}rvi, T.~Kaukoranta, O.~Nevalainen. 1997.
\newblock {Genetic algorithms for large-scale clustering problems}.
\newblock {\it The Computer Journal\/} {\bf 40}(9) 547--554.

\bibitem[{Fr{\"{a}}nti et~al.(2014)Fr{\"{a}}nti, Rezaei, and Zhao}]{Franti2014}
Fr{\"{a}}nti, P., M.~Rezaei, Q.~Zhao. 2014.
\newblock {Centroid index: Cluster level similarity measure}.
\newblock {\it Pattern Recognition\/} {\bf 47}(9) 3034--3045.

\bibitem[{Hamerly(2010)}]{Hamerly2010}
Hamerly, G. 2010.
\newblock {Making k-means even faster}.
\newblock {\it SDM'10, SIAM International Conference on Data Mining\/}.
  130--140.

\bibitem[{Han et~al.(2011)Han, Kamber, and Pei}]{Han2011}
Han, J., M.~Kamber, J.~Pei. 2011.
\newblock {\it {Data mining: concepts and techniques}\/}.
\newblock 3rd ed. Morgan Kaufmann.

\bibitem[{Hansen and Mladenovi{\'{c}}(2001)}]{Hansen2001a}
Hansen, P., N.~Mladenovi{\'{c}}. 2001.
\newblock {J-Means: A new local search heuristic for minimum sum of squares
  clustering}.
\newblock {\it Pattern Recognition\/} {\bf 34}(2) 405--413.

\bibitem[{Hartigan and Wong(1979)}]{Hartigan1979}
Hartigan, J.A., M.A. Wong. 1979.
\newblock {Algorithm AS 136: A k-means clustering algorithm}.
\newblock {\it Applied Statistics\/} {\bf 28}(1) 100--108.

\bibitem[{Hoffer et~al.(2017)Hoffer, Hubara, and Soudry}]{Hoffer2017}
Hoffer, E., I.~Hubara, D.~Soudry. 2017.
\newblock {Train longer, generalize better: Closing the generalization gap in
  large batch training of neural networks}.
\newblock {\it Advances in Neural Information Processing Systems\/}.
  1729--1739.

\bibitem[{Holland(1975)}]{Holland1975}
Holland, J.H. 1975.
\newblock {\it {Adaptation in Natural and Artificial Systems}\/}.
\newblock The University of Michigan Press, Ann Arbor, MI.

\bibitem[{Hruschka et~al.(2009)Hruschka, Campello, Freitas, and
  de~Carvalho}]{Hruschka2009}
Hruschka, E.R., R.J.G.B. Campello, A.A. Freitas, A.C.P.L.F. de~Carvalho. 2009.
\newblock {A survey of evolutionary algorithms for clustering}.
\newblock {\it IEEE Transactions on Systems, Man and Cybernetics Part C:
  Applications and Reviews\/} {\bf 39}(2) 133--155.

\bibitem[{Hubert and Arabie(1985)}]{Hubert1985}
Hubert, L., P.~Arabie. 1985.
\newblock {Comparing partitions}.
\newblock {\it Journal of Classification\/} {\bf 2}(1) 193--218.

\bibitem[{Ismkhan(2018)}]{Ismkhan2018}
Ismkhan, H. 2018.
\newblock {I-k-means-+: An iterative clustering algorithm based on an enhanced
  version of the k-means}.
\newblock {\it Pattern Recognition\/} {\bf 79}(1) 402--413.

\bibitem[{Jain(2010)}]{Jain2010}
Jain, A.K. 2010.
\newblock {Data clustering: 50 years beyond K-means}.
\newblock {\it Pattern Recognition Letters\/} {\bf 31}(8) 651--666.

\bibitem[{Karmitsa et~al.(2017)Karmitsa, Bagirov, and Taheri}]{Karmitsa2017}
Karmitsa, N., A.M. Bagirov, S.~Taheri. 2017.
\newblock {New diagonal bundle method for clustering problems in large data
  sets}.
\newblock {\it European Journal of Operational Research\/} {\bf 263}(2)
  367--379.

\bibitem[{Karmitsa et~al.(2018)Karmitsa, Bagirov, and Taheri}]{Karmitsa2018}
Karmitsa, N., A.M. Bagirov, S.~Taheri. 2018.
\newblock Clustering in large data sets with the limited memory bundle method.
\newblock {\it Pattern Recognition\/} {\bf 83} 245--259.

\bibitem[{Keskar et~al.(2017)Keskar, Mudigere, Nocedal, Smelyanskiy, and
  Tang}]{Keskar2017}
Keskar, N.S., D.~Mudigere, J.~Nocedal, M.~Smelyanskiy, P.T.P. Tang. 2017.
\newblock {On large-batch training for deep learning: Generalization gap and
  sharp minima}.
\newblock {\it ICLR'17, International Conference on Learning
  Representations\/}.

\bibitem[{Kivij{\"{a}}rvi et~al.(2003)Kivij{\"{a}}rvi, Fr{\"{a}}nti, and
  Nevalainen}]{Kivijarvi2003}
Kivij{\"{a}}rvi, J., P.~Fr{\"{a}}nti, O.~Nevalainen. 2003.
\newblock {Self-adaptive genetic algorithm for clustering}.
\newblock {\it Journal of Heuristics\/} {\bf 9}(2) 113--129.

\bibitem[{Krishna and Murty(1999)}]{Krishna1999}
Krishna, K., M.N. Murty. 1999.
\newblock {Genetic k-means algorithm}.
\newblock {\it IEEE Transactions on Systems, Man, and Cybernetics, Part B:
  Cybernetics\/} {\bf 29}(3) 433--439.

\bibitem[{Kuhn(1955)}]{Kuhn1955}
Kuhn, H.W. 1955.
\newblock {The Hungarian method for the assignment problem}.
\newblock {\it Naval Research Logistics\/} {\bf 2}(1-2) 83--97.

\bibitem[{Kvalseth(1987)}]{Kvalseth1987}
Kvalseth, T.O. 1987.
\newblock {Entropy and correlation: Some comments}.
\newblock {\it IEEE Transactions on Systems, Man and Cybernetics\/} {\bf 17}(3)
  517--519.

\bibitem[{Lecun et~al.(2015)Lecun, Bengio, and Hinton}]{Lecun2015}
Lecun, Y., Y.~Bengio, G.~Hinton. 2015.
\newblock {Deep learning}.
\newblock {\it Nature\/} {\bf 521} 436--444.

\bibitem[{Likas et~al.(2003)Likas, Vlassis, and Verbeek}]{Likas2003}
Likas, A., N.~Vlassis, J.J. Verbeek. 2003.
\newblock {The global k-means clustering algorithm}.
\newblock {\it Pattern Recognition\/} {\bf 36}(2) 451--461.

\bibitem[{Lloyd(1982)}]{Lloyd1982}
Lloyd, S.P. 1982.
\newblock {Least squares quantization in PCM}.
\newblock {\it IEEE Transactions on Information Theory\/} {\bf 28}(2) 129--137.

\bibitem[{Lu et~al.(2004)Lu, Lu, Fotouhi, Deng, and Brown}]{Lu2004a}
Lu, Y., S.~Lu, F.~Fotouhi, Y.~Deng, S.J. Brown. 2004.
\newblock {FGKA: A fast genetic K-means clustering algorithm}.
\newblock {\it Proceedings of the 2004 ACM Symposium on Applied Computing\/}
  622--623.

\bibitem[{Maulik and Bandyopadhyay(2000)}]{Maulik2000}
Maulik, U., S.~Bandyopadhyay. 2000.
\newblock {Genetic algorithm-based clustering technique}.
\newblock {\it Pattern Recognition\/} {\bf 33} 1455--1465.

\bibitem[{Merz and Zell(2002)}]{Merz2002}
Merz, P., A.~Zell. 2002.
\newblock {Clustering gene expression profiles with memetic algorithms}.
\newblock {\it Parallel Problem Solving from Nature\/}. Springer, 811--820.

\bibitem[{Ordin and Bagirov(2015)}]{Ordin2015}
Ordin, B., A.M. Bagirov. 2015.
\newblock {A heuristic algorithm for solving the minimum sum-of-squares
  clustering problems}.
\newblock {\it Journal of Global Optimization\/} {\bf 61}(2) 341--361.

\bibitem[{Sarkar et~al.(1997)Sarkar, Yegnanarayana, and Khemani}]{Sarkar1997}
Sarkar, M., B.~Yegnanarayana, D.~Khemani. 1997.
\newblock {A clustering algorithm using an evolutionary programming-based
  approach}.
\newblock {\it Pattern Recognition Letters\/} {\bf 18}(10) 975--986.

\bibitem[{Scheunders(1997)}]{Scheunders1997a}
Scheunders, P. 1997.
\newblock {A comparison of clustering algorithms applied to color image
  quantization}.
\newblock {\it Pattern Recognition Letters\/} {\bf 18} 1379--1384.

\bibitem[{Selim and Alsultan(1991)}]{Selim1991}
Selim, S.Z., K.~Alsultan. 1991.
\newblock {A simulated annealing algorithm for the clustering problem}.
\newblock {\it Pattern Recognition\/} {\bf 24}(10) 1003--1008.

\bibitem[{S{\"{o}}rensen and Sevaux(2006)}]{Soerensen2006}
S{\"{o}}rensen, K., M.~Sevaux. 2006.
\newblock {MAPM: Memetic algorithms with population management}.
\newblock {\it Computers {\&} Operations Research\/} {\bf 33}(5) 1214--1225.

\bibitem[{Steinley(2006)}]{Steinley2006}
Steinley, D. 2006.
\newblock {K-means clustering: A half-century synthesis}.
\newblock {\it British Journal of Mathematical and Statistical Psychology\/}
  {\bf 59}(1) 1--34.

\bibitem[{Sugar and James(2003)}]{Sugar2003}
Sugar, C.A., G.M. James. 2003.
\newblock {Finding the number of clusters in a dataset}.
\newblock {\it Journal of the American Statistical Association\/} {\bf 98}(463)
  750--763.

\bibitem[{Vidal et~al.(2012)Vidal, Crainic, Gendreau, Lahrichi, and
  Rei}]{Vidal2012}
Vidal, T., T.G. Crainic, M.~Gendreau, N.~Lahrichi, W.~Rei. 2012.
\newblock {A hybrid genetic algorithm for multidepot and periodic vehicle
  routing problems}.
\newblock {\it Operations Research\/} {\bf 60}(3) 611--624.

\end{thebibliography}
\end{document}